\documentclass[sigconf, nonacm]{aamas}

\usepackage{balance} 

\usepackage[disable, colorinlistoftodos, color=blue!20!white]{todonotes} 
\usepackage{url}
\usepackage{algorithm}
\usepackage{caption}
\usepackage{subcaption}
\usepackage[noend]{algpseudocode}
\algrenewcommand\algorithmicrequire{\textbf{Input:}}
\algrenewcommand\algorithmicensure{\textbf{Output:}}
\usepackage{amsmath}
\usepackage{amsfonts}
\usepackage{booktabs}

\usepackage{tikz}
\usepackage{adjustbox}
\usetikzlibrary{shapes.geometric, arrows.meta, positioning,fit,backgrounds,calc}

\tikzstyle{process} = [rectangle, rounded corners, minimum width=2cm, minimum height=0.5cm,text centered, draw=black, font=\small, fill=blue!10, text width=2.3cm]
\tikzstyle{arrow} = [thick,->,>=stealth]
\tikzstyle{startstop} = [ellipse, minimum width=2cm, minimum height=0.5cm, text centered, draw=black, font=\footnotesize, fill=red!10]

\newcommand{\donetodo}[1]{%
  \todo[
    color=black!5,
    bordercolor=black!15,   
    textcolor=black!25
  ]{#1}%
}

\newcommand{\review}[1]{%
    \todo[
        color=orange!10,
        bordercolor=orange!40!black!40,
        textcolor=orange!60!black!100,
    ]{#1}%
}

\definecolor{planblue}{HTML}{DCEAF7}
\definecolor{nlpgreen}{HTML}{DDEED5}
\definecolor{lcsorange}{HTML}{F9E4C3}
\definecolor{validpurple}{HTML}{E7DDF2}
\definecolor{recoverred}{HTML}{F5D6D2}
\definecolor{graybox}{HTML}{F2F2F2}
\definecolor{darkgray}{HTML}{4A4A4A}

\makeatletter
\gdef\@copyrightpermission{
  \begin{minipage}{0.2\columnwidth}
   \href{https://creativecommons.org/licenses/by/4.0/}{\includegraphics[width=0.90\textwidth]{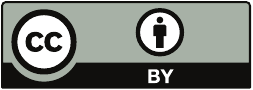}}
  \end{minipage}\hfill
  \begin{minipage}{0.8\columnwidth}
   \href{https://creativecommons.org/licenses/by/4.0/}{This work is licensed under a Creative Commons Attribution International 4.0 License.}
  \end{minipage}
  \vspace{5pt}
}
\makeatother

\renewcommand\footnotetextcopyrightpermission[1]{}

\setcopyright{ifaamas}
\acmConference[AAMAS 2027]{Proc.\@ of the 26th International Conference
on Autonomous Agents and Multiagent Systems (AAMAS 2027)}{3--7 May 2027}{Hanoi, Vietnam}{M.~Baldoni, F.~Fang, W.~Yeoh, N.~Yorke-Smith (eds.)}
\copyrightyear{2027}
\acmYear{2027}
\acmDOI{}
\acmISBN{}

\acmSubmissionID{}

\submissionType{}

\title{How Far Are LLMs from Symbolic Planners? \\ An NLP-Based Perspective}

\author{
    Ma'ayan Armony
}
\affiliation{
  \institution{Department of Informatics}
  \city{King's College London}
  \country{UK}
}

\email{maayan.armony@kcl.ac.uk}
\author{
    Albert Meroño-Peñuela
}
\affiliation{
  \institution{Department of Informatics}
  \city{King's College London}
  \country{UK}
}
\email{albert.merono@kcl.ac.uk}
\author{
    Gerard Canal
}
\affiliation{
  \institution{Department of Informatics}
  \city{King's College London}
  \country{UK}
}
\email{gerard.canal@kcl.ac.uk}

\begin{abstract}
AI planning provides agents the ability to act autonomously towards achieving a goal in a given environment. Recent advances in Large Language Models (LLMs) have introduced the prospect of using unstructured input for planning, making LLMs in planning an active area of research, especially for supporting autonomous agents. 
However, LLMs remain unreliable as long-horizon planners, since some of the plans they produce contain mistaken or hallucinated actions; executing such plans could lead to unexpected and harmful outcomes.
Existing benchmarks for LLMs in planning primarily rely on binary evaluation metrics, providing limited insight into the quality of plans that do not achieve the goal, and how far those are from becoming valid.
In this paper, we frame the evaluation of LLMs in planning as a natural language processing (NLP) task, as LLMs are themselves NLP models. We propose a pipeline comprising three NLP-based steps that assess plan quality from complementary perspectives, followed by a symbolic planning step to achieve a valid plan, to provide a finer-grained understanding of the quality of invalid plans.
In our experiments, we find no clear evidence of logical derivation or state-tracking during LLM plan generation --
on average, for all non-empty plans, only 3.55 actions per plan are executable, despite symbolically-generated optimal plans having an average length of 9.47 actions, which suggests that LLMs are still far from the reliability of symbolic planners. Our suite of metrics is further used to investigate patterns in LLM plan generation over time, and can provide a useful metric for assessing the quality of individual invalid plans. 
\end{abstract}


\newcommand{\BibTeX}{\rm B\kern-.05em{\sc i\kern-.025em b}\kern-.08em\TeX}

\begin{document}




\maketitle 


\section{Introduction} 
AI planning enables agents to find plans that achieve a specified goal in a given environment.
Symbolic planners rely on formal descriptions of the applicable actions, their preconditions and effects, the existing objects, and the initial and goal states; these are often expressed using the Planning Domain Definition Language (PDDL)~\cite{mcdermott1998pddl}.
To allow more expressive and flexible task descriptions, Large Language Models (LLMs) \cite{vaswani2017attention,devlin2019bert,brown2020language} have been increasingly utilised in planning \cite{singh2023progprompt,silver2024generalized,2024chenautotamp}. LLMs are deep neural networks trained on large corpora to generate token sequences based on input prompts. Their outputs are derived through statistical computation rather than logical derivation.
As a consequence of their architecture and design, which aim to reliably predict the next token, LLMs do not evaluate the constraints and effects of actions. Such evaluation is the backbone of symbolic planners, enabling them to produce sound and consistent plans. Despite recent advancements, LLMs still struggle to demonstrate accurate and reliable outcomes \cite{valmeekam2023planbench,huang2025chasing}. 

\citet{lal2024cat} found that LLMs barely outperform random choice (50\%) at predicting action order, and their output is inconsistent with logical assumptions. They also found that reordering independent actions incorrectly alters the model's response. 
This suggests that planning with LLMs is, at least in part, a Natural Language Processing (NLP) task, where the model is heavily influenced by input structure rather than relying solely on the underlying logic it encodes.
Several evaluation strategies have been proposed for LLM planning~\cite{liu2023llm+p,valmeekam2023planbench,silver2022pddl}. 
The most used metric for evaluating LLM-generated plans is success rate (SR) -- the proportion of plans that reach the goal while adhering to domain constraints -- for a given model, prompt type, and domain. 
While SR offers some insight into the quality of the generation strategy, it reduces evaluation to a binary classification, disregarding the quality of invalid plans. 
Other approaches assess LLMs on a set of planning tasks, such as plan validation or reasoning over state transitions. However, these typically address isolated capabilities, overlooking LLMs' generative behaviour, whose output patterns may provide clues about the underlying process. Their metrics, such as SR, precision, recall, or F1-score, do not capture how far the failing plans are from success, nor whether the LLM reasons over the input. \citet{huang2025chasing} compute plans' executability rate~(ER) as the proportion of plans following domain constraints, but ER remains plan-level, overlooking action-level quality. 
Using a binary metric on a small set of plans for a specific task provides a coarse and unstable evaluation, with all invalid and non-executable plans being indistinguishable.

In this paper, with the aim of evaluating LLM-generated plans, we reposition the task of planning with LLMs also as an NLP task, treating plans as a form of language output during their evaluation. This allows us to investigate patterns that planning tools may not reflect, and are essential to understanding LLMs' behaviour in planning. While existing methods are grounded in symbolic planning, focusing only on aspects such as validity or executability, we first assess plan structural and semantic proximity to a valid plan. This shift allows us to analyse model behaviour at the token and action levels, leading to a more nuanced observation of linguistic and structural patterns in the output, beyond whether it satisfies classical planning objectives. 
We aim to answer the following questions:
\begin{enumerate}
    \item[\textbf{RQ1.}] How far are LLMs from producing valid plans? 
    \item[\textbf{RQ2.}] Is SR a sufficient metric for representing plan quality?
    \item[\textbf{RQ3.}] Is there evidence of logical derivation and state-tracking in LLM plan generation?
\end{enumerate}
We contribute a new evaluation pipeline with (a) NLP- and planning-based deterministic metrics, which can better capture the nature of LLM-generated plans, and support distinct evaluation of individual invalid plans, and (b) a mechanism combining NLP-based plan manipulation with symbolic validation and replanning to analyse LLM planning behaviour. \donetodo{Remove the notion of plan recovery}
Using this pipeline, we show that: (1) most invalid plans (58.3\%) violate domain constraints already by the first action, pointing to a lack of state-tracking required in planning;
(2) even invalid LLM plans that follow domain constraints rarely advance towards the goal, and (3) NLP-refinement is much less effective on current LLMs' plans than with 2025 LLMs, revealing changes in generation patterns that cannot be captured solely with planning-based and binary evaluations. 
The implementation of the pipeline can be found on GitHub\footnote{\url{https://github.com/maayan25/llm-plan-evaluation}}.  
\section{Related Work}
\paragraph{LLM Reasoning.}
Prior work has examined LLM reasoning abilities, including causal \cite{zhang-etal-2023-causal} and temporal reasoning \cite{tan2023towards}. 
LLMs have been shown to struggle with reasoning tasks and often fail to use external feedback for self-correction \cite{huang2023large,illusion-of-thinking}. 
Prompting strategies such as Chain of Thought (CoT)~\cite{wei2022chain} and least-to-most~\cite{zhou2022least} elicit advanced reasoning by decomposing tasks into intermediate steps.

\paragraph{LLMs in planning.}
Recent work has explored the integration of LLMs into planning. A variety of strategies have emerged, ranging from multi-stage generation pipelines, such as environmental feedback after each action execution \cite{huang2023inner}, to using CoT prompting to improve reasoning \cite{silver2024generalized,yao2023react}.
As LLMs are trained on textual corpora, some methods suggest planning in natural language (NL) rather than PDDL \cite{song2023llm,brohan2023can}. Others synthesise plans as programs~\cite{singh2023progprompt}.
To overcome reasoning limitations, hybrid approaches use LLMs to support symbolic planners, for instance, by translating NL into structured inputs \cite{2024chenautotamp,liu2023llm+p}, injecting domain knowledge \cite{dagan2023dynamic}, producing high-level plans \cite{song2023llm}, or generating sub-goals \cite{silvia2024plancollabnl}. An emerging area of research leverages LLMs to create programs that can generate plans or heuristic functions, as the programs themselves can be verified for their soundness and completeness \cite{katz2024thought,correa2026classical}.
The diversity of these strategies underscores the need for a comprehensive evaluation to assess plan quality and analyse the models' reasoning capabilities when it assumes the role of a planner. Most works evaluate the quality of generated plans through task success rates, overlooking how close failed attempts are to being successful -- a crucial aspect for understanding and improving model behaviour. 

\paragraph{Evaluation of LLMs in planning.}
Several benchmarks have been proposed for evaluating LLMs in planning.
PlanBench~\cite{valmeekam2023planbench} evaluates LLM-based planning across two domain families using SR for various planning tasks with different prompting strategies, including NL, PDDL, and CoT NL for most domains. 
Subsequent analysis~\cite{valmeekam2023planning} shows that finetuning yields only limited improvements in SR, and that obfuscating the domain degrades performance, suggesting a reliance on prompt semantics.
The authors further claim that the symbolic planner LPG~\cite{gerevini2002lpg} can repair LLM-generated plans more efficiently than random plans, but repair distance is not incorporated as a metric within the benchmark itself. TravelPlanner~\cite{xie2024travelplanner} limits the number of planning steps and reports the portion of satisfied constraints.
\citet{silver2022pddl} validate actions throughout generation. They report that mistaken actions are common and that, in many domains, SR is 0\% even with validation. To assess whether LLMs can guide symbolic planners, they use the number of nodes created and expanded during search as additional metrics, demonstrating improvements over a pure planning baseline. 
While LLM-generated plans are shown to be better for guiding search than random plans, they remain less efficient than symbolic planners, and the extent of required intervention during validation is not explicitly accounted for, making it hard to infer the quality of LLM-generated plans.
\citet{huang2025chasing} address the limitations of SR as a sole metric by introducing the executability rate (ER) as an additional measure. They find that CoT prompting improves ER but not SR, and use reinforcement learning with a reward based on the Longest Contiguous Common Subsequence (LCS) to improve both metrics. However, ER remains a binary, plan-level evaluation.

Other benchmarks focus on isolated reasoning and planning abilities.
ACPBench~\cite{kokel2024acpbench} evaluates decision-making across multiple planning tasks and finds that smaller fine-tuned models can outperform larger ones, reinforcing the view that LLMs often imitate rather than reason.
ACPBench Hard~\cite{kokel2025acpbench} extends this to open-ended decision-making questions on which models achieve less than 60\% SR. 
ActPlan-1k~\cite{su-etal-2024-actplan} uses more granular metrics, including LCS and a model fine-tuned for plan quality evaluation. However, this model does not fully align with human evaluation, which could obscure the sources of errors when used as a quality metric.
FlowBench~\cite{xiao2024flowbench} and CAT-BENCH~\cite{lal2024cat} focus on workflow planning and temporal reasoning, respectively, using metrics such as precision, recall, and temporal consistency. FlowBench's scoring and failure analysis rely on GPT-4, raising concerns about the evaluation's robustness and reliability. 
Overall, existing evaluation approaches examine LLM-based planning from multiple perspectives but mostly rely on binary, plan-level metrics such as task success or validity, even when intermediate planning steps are considered \cite{wei2025plangenllms,Cao2025LargeLM}. As argued by~\citet{illusion-of-thinking}, such metrics provide limited insight into underlying reasoning capabilities, highlighting the need for methods that capture how far plans are from valid solutions.

\section{Methodology}
Our evaluation pipeline consists of three parts: similarity of LLM-generated plans to a symbolically-generated plan, plan manipulation using NLP transformations to investigate the structure and semantics of LLM-generated plans,
and a final refinement strategy for additional analysis of how far LLMs are from symbolic planners. The refinement, based on a symbolic plan, is used to examine the quality of plans from various angles and gain a better understanding of the plan's quality, and not as a repair strategy. The pipeline is illustrated in Figure \ref{fig:plan_pipeline}, and the introduced metrics are shown in Figure \ref{fig:plan_pipeline_metrics}, with a summary table of all metrics in the Appendix.

\subsection{LLM Plan Generation}
We utilise the PlanBench~\cite{valmeekam2023planbench} pipeline for prompt generation, plan generation, and plan parsing. The PlanBench framework is suitable for our study, as it enables structured interaction with LLMs with various prompt types, whilst allowing for PDDL-based inputs and outputs. 
We extend PlanBench with additional LLMs, and add prompts and parsing for new and more complex PDDL domains. We maintain a deterministic and rule-based parsing, as included in the latest PlanBench version, to limit the source of LLM errors to the generation itself. 
To focus on the model's generation behavior rather than on the syntax correctness of the output, we allow further flexibility for differing LLM behaviours, by accepting various output formats. For example, \texttt{t0} and \texttt{truck0} are both accepted as an object name in Logistics, and automatically parsed as the object's PDDL name (\texttt{t0}). 
PlanBench uses the Fast-Downward planner (FD) \cite{helmert2006fast} to generate optimal and valid ground-truth (GT) plans.
For each instance, we denote the optimal GT plan as $\pi^*$ and the LLM-generated plan as $\pi_{\text{LLM}}$. LLM-generated plans that successfully reach the goal while adhering to the domain dynamics are marked as valid. We define SR as the portion of valid plans across all problem instances. 

Comparing to a single GT plan can be overly strict on LLM plan structure.
Therefore, we run an additional planner, SymK \cite{speck-et-al-aaai2020}, to get top-k plans for each problem instance. We propose a heuristic to choose the plan used in the evaluation pipeline, denoted $\pi_\text{GT}$, as described in Section \ref{sect:choosing-gt}.
Our pipeline yields four intermediate plans:
$\pi_{\text{NLP}}$ is an NLP-manipulated plan that is the most similar to $\pi_{\text{GT}}$, as detailed in Section \ref{nlp-manipulation}; the best portions of $\pi_{\text{LLM}}$ and $\pi_{\text{NLP}}$ are set as $\pi_{\text{LCS}}$ and $\pi_{\text{NLP-LCS}}$, respectively, as described in Section \ref{best-lcs}; and lastly a plan $\pi_{\text{VAL}}$ which is completed with a symbolic planner, as detailed in Section \ref{sect:completion}.

\paragraph{Running example} To demonstrate the methodology on an invalid plan, we use a running example from the Blocksworld domain (3-blocks version, see description in \ref{sect:tasks-domains}) \cite{gupta1992complexity} with the NL one-shot prompt-type, problem instance 100, and Gemma 4 26B. Figure \ref{fig:blocksworld-problem} illustrates the initial and goal states for this problem.
The example is used throughout the section, and a consolidated table of the results can be found in the Appendix. $\pi_{\text{LLM}}$, $\pi_{\text{GT}}$, and $\pi^*$ are: 

\begin{table}[h]
  \centering
  \footnotesize
  \begin{adjustbox}{max width=0.7\textwidth}
  \begin{tabular}{r@{\ }l@{\quad}r@{\ }l@{\quad}r@{\ }l}
    \multicolumn{2}{c}{\textbf{\small$\pi_{\text{LLM}}$}} & \multicolumn{2}{c}{\textbf{\small$\pi_{\text{GT}}$}} & \multicolumn{2}{c}{\textbf{\small$\pi^*$}}\\
    \midrule
    1 & \texttt{(pick-up c)} & 1 & \texttt{(unstack c a)} & 1 & \texttt{(pick-up b)} \\
    2 & \texttt{(stack c a)} & 2 & \texttt{(stack c a)} & 2 & \texttt{(stack b c)} \\
    3 & \texttt{(stack b c)} & 3 & \texttt{(pick-up b)} &  &  \\
    4 & \texttt{(stack a b)} & 4 & \texttt{(stack b c)} &  &  \\
  \end{tabular}
  \end{adjustbox}
\end{table}

\begin{figure}[tbp]
\centering
\begin{adjustbox}{max width=0.5\textwidth}
\begin{tikzpicture}[
    >=Latex,
    node distance=0.6cm and 0.6cm
]

\node[
    draw=darkgray,
    thick,
    rounded corners=2pt,
    fill=white,
    minimum height=1.15cm,
    minimum width=2.0cm,
    align=center,
    font=\sffamily
] (plan) {
    LLM Plan\\
    Generation
};

\node[
    draw=darkgray,
    thick,
    rounded corners=2pt,
    fill=white,
    minimum height=1.15cm,
    minimum width=2.0cm,
    align=center,
    font=\sffamily,
    below=of plan
] (fd) {
    FD planner
};

\node[
    draw=darkgray,
    thick,
    rounded corners=2pt,
    fill=graybox,
    minimum height=1.15cm,
    minimum width=2.0cm,
    align=center,
    font=\sffamily,
    right=of plan
] (gt) {
    A. Choosing GT \\
    Plan
};

\node[
    draw=darkgray,
    thick,
    rounded corners=2pt,
    fill=planblue,
    minimum height=1.15cm,
    minimum width=2.0cm,
    align=center,
    font=\sffamily,
    right=of gt
] (eval) {
    B. Plan Evaluation\\
    and Validation
};

\node[
    draw=darkgray,
    thick,
    rounded corners=2pt,
    fill=nlpgreen,
    minimum height=1.15cm,
    minimum width=2.0cm,
    align=center,
    font=\sffamily,
    right=of eval
] (nlp) {
    C. NLP\\
    Manipulation
};

\node[
    draw=darkgray,
    thick,
    rounded corners=2pt,
    fill=lcsorange,
    minimum height=1.15cm,
    minimum width=2.0cm,
    align=center,
    font=\sffamily,
    right=of nlp
] (lcs) {
    D. Best LCS\\
    Selection
};

\node[
    draw=darkgray,
    thick,
    rounded corners=2pt,
    fill=recoverred,
    minimum height=1.15cm,
    minimum width=2.0cm,
    align=center,
    font=\sffamily,
    below=1.5cm of eval
] (recovery) {
    E. Plan Refinement\\
    / Replanning
};


\node[
    draw=darkgray,
    thick,
    rounded corners=2pt,
    fill=white,
    minimum width=1.3cm,
    minimum height=0.7cm,
    align=center,
    font=\sffamily,
    above=0.8cm of gt
] (pigt) {$\pi_{\text{GT}}$};

\node[
    draw=darkgray,
    thick,
    rounded corners=2pt,
    fill=white,
    minimum width=1.3cm,
    minimum height=0.7cm,
    align=center,
    font=\sffamily,
    above=0.8cm of plan
] (pi0) {$\pi_{\text{LLM}}$};

\node[
    draw=darkgray,
    thick,
    rounded corners=2pt,
    fill=white,
    minimum width=1.3cm,
    minimum height=0.7cm,
    align=center,
    font=\sffamily,
    below=0.8cm of fd
] (pistar) {$\pi^*$};

\node[
    draw=darkgray,
    thick,
    rounded corners=2pt,
    fill=white,
    minimum width=1.3cm,
    minimum height=0.7cm,
    align=center,
    font=\sffamily,
    above=0.8cm of nlp
] (pi1) {$\pi_{\text{NLP}}$};

\node[
    draw=darkgray,
    thick,
    rounded corners=2pt,
    fill=white,
    minimum width=1.3cm,
    minimum height=0.7cm,
    align=center,
    font=\sffamily,
    above=0.8cm of lcs
] (pi23) {$\pi_{\text{LCS}},\pi_{\text{NLP-LCS}}$};

\node[
    draw=darkgray,
    thick,
    rounded corners=2pt,
    fill=white,
    minimum width=1.3cm,
    minimum height=0.7cm,
    align=center,
    font=\sffamily,
    below=0.8cm of recovery
] (pi4) {$\pi_{\text{VAL}}$};

\draw[
    ->,
    thick,
    draw=darkgray
] (plan) -- (gt);

\draw[
    ->,
    thick,
    draw=darkgray
] (gt) -- (eval);

\draw[
    ->,
    thick,
    draw=darkgray
] (eval) -- (nlp);

\draw[
    ->,
    dashed,
    draw=darkgray
] (nlp.west)
    -- ++(0,0.9)
    -| (eval.north);

\draw[
    ->,
    thick,
    draw=darkgray
] (nlp) -- (lcs);

\draw[
    ->,
    dashed,
    draw=darkgray
] (lcs.west)
    -- ++(0,1.2)
    -| (eval.north);


\draw[
    ->,
    thick,
    draw=darkgray
] (gt) -- (pigt);

\draw[
    ->,
    thick,
    draw=darkgray
] (plan) -- (pi0);

\draw[
    ->,
    thick,
    draw=darkgray
] (fd) -- (pistar);

\draw[
    ->,
    thick,
    draw=darkgray
] (nlp) -- (pi1);

\draw[
    ->,
    thick,
    draw=darkgray
] (lcs) -- (pi23);

\draw[
    ->,
    thick,
    draw=darkgray
] (recovery) -- (pi4);

\draw[
    ->,
    thick,
    draw=darkgray
] (eval.south)
    -- node[
        left,
        font=\sffamily\small\itshape,
        text=darkgray
    ] {invalid $\pi_{\text{LLM}}$}
    (recovery.north);
\end{tikzpicture}
\end{adjustbox}
\caption{Overview of the pipeline - intermediate plans}
\label{fig:plan_pipeline}
\end{figure}

\begin{figure}[tbp]
\centering
\begin{tikzpicture}[
  block/.style={draw, minimum width=0.6cm, minimum height=0.45cm},
  node distance=0.0cm
]

\node (init) {\footnotesize Initial};

\node[block, below right=0.25cm and 0.25cm of init] (b1) {b};
\node[block, right=0.25cm of b1] (a1) {a};
\node[block, above=of a1] (c1) {c};

\draw (-0.5,-0.9) -- (2.5,-0.9);

\node[right=2.5cm of init] (goal) {\footnotesize Goal};

\node[block, below right=0.25cm and 0.25cm of goal] (a2) {a};
\node[block, above=of a2] (c2) {c};
\node[block, above=of c2] (b2) {b};

\draw (3.1,-0.9) -- (5.8,-0.9);

\end{tikzpicture}
\caption{BW 3 problem for running example - instance 100}
\label{fig:blocksworld-problem}
\end{figure}

\subsection{Choosing a Ground Truth Plan} \label{sect:choosing-gt}
We use the SymK planner \cite{speck-et-al-aaai2020}, an extension of Fast-Downward, to generate top-k plans, i.e. plans with the lowest cost, for each problem instance ($k=5$). We propose a heuristic for choosing a $\pi_\text{GT}$ for each $\pi_\text{LLM}$, based on structural similarity between the two plans. This heuristic is a simplified version of the full plan evaluation module detailed in Section \ref{sect:plan_evaluation}. 
For each candidate $\pi_{\text{GT}_n}$ ($1 \le n \le k$) we compute a score $S_n$, as defined at the end of Section \ref{sect:plan_evaluation}. 
The candidate with the highest $S_n$ becomes $\pi_\text{GT}$. 

\subsection{Plan evaluation} \label{sect:plan_evaluation}
\begin{figure}[tbp]
\centering
\begin{adjustbox}{max width=0.5\textwidth}
\begin{tikzpicture}[
    >=Latex,
    node distance=0.75cm
]

\definecolor{graybox}{HTML}{F2F2F2}
\definecolor{planblue}{HTML}{DCEAF7}
\definecolor{nlpgreen}{HTML}{DDEED5}
\definecolor{lcsorange}{HTML}{F9E4C3}
\definecolor{validpurple}{HTML}{E7DDF2}
\definecolor{darkgray}{HTML}{4A4A4A}

\node[
    draw=darkgray,
    thick,
    rounded corners=2pt,
    fill=planblue,
    minimum height=0.8cm,
    minimum width=2.2cm,
    align=center,
    font=\sffamily\small
] (val) {
    Validate\\
    (VAL)
};

\node[
    draw=darkgray,
    thick,
    rounded corners=2pt,
    fill=planblue,
    minimum height=0.8cm,
    minimum width=2.2cm,
    align=center,
    font=\sffamily\small,
    below=of val
] (sim) {
    Simulate
};

\node[
    draw=darkgray,
    thick,
    rounded corners=2pt,
    fill=white,
    minimum height=0.8cm,
    minimum width=2.2cm,
    align=center,
    font=\sffamily\small,
    right=0.9cm of val
] (lea) {
    LEA
};


\node[
    draw=darkgray,
    thick,
    rounded corners=2pt,
    fill=lcsorange,
    minimum height=0.8cm,
    minimum width=2.0cm,
    align=center,
    font=\sffamily\small,
    below=1cm of sim
] (lcs) {
    LCS
};

\node[
    draw=darkgray,
    thick,
    rounded corners=2pt,
    fill=lcsorange,
    minimum height=0.8cm,
    minimum width=2.0cm,
    align=center,
    font=\sffamily\small,
    right=1cm of lcs
] (penalty) {
    Length\\
    factor
};

\node[
    draw=darkgray,
    thick,
    rounded corners=2pt,
    fill=nlpgreen,
    minimum height=0.8cm,
    minimum width=2.5cm,
    align=center,
    font=\sffamily\small,
    right=1cm of lea
] (correct) {
    Correct actions
};

\node[
    draw=darkgray,
    thick,
    rounded corners=2pt,
    fill=nlpgreen,
    minimum height=0.8cm,
    minimum width=2.5cm,
    align=center,
    font=\sffamily\small,
    below=of correct
] (misplaced) {
    Misplaced actions
};

\node[
    draw=darkgray,
    thick,
    rounded corners=2pt,
    fill=nlpgreen,
    minimum height=0.8cm,
    minimum width=2.5cm,
    align=center,
    font=\sffamily\small,
    below=of misplaced
] (pairing) {
    Partial\\ action similarity
};

\node[
    draw=darkgray,
    thick,
    rounded corners=2pt,
    fill=nlpgreen,
    minimum height=0.8cm,
    minimum width=2.5cm,
    align=center,
    font=\sffamily\small,
    below=of pairing
] (redundant) {
    Evaluate\\
    redundant actions
};

\node[
    draw=darkgray,
    thick,
    rounded corners=2pt,
    fill=validpurple,
    minimum height=0.8cm,
    minimum width=2.5cm,
    align=center,
    font=\sffamily\small,
    below=of redundant
] (stv) {
    Steps\\ to Validity
};

\node[
    draw=darkgray,
    thick,
    rounded corners=2pt,
    fill=white,
    minimum width=1.5cm,
    minimum height=0.65cm,
    align=center,
    font=\sffamily\small,
    right=0.7cm of pairing
] (aqm) {
    AQM
};

\node[
    draw=darkgray,
    thick,
    rounded corners=2pt,
    fill=white,
    minimum width=1.5cm,
    minimum height=0.65cm,
    align=center,
    font=\sffamily\small,
    right=0.7cm of redundant
] (npaqm) {
    NP-AQM
};

\node[
    draw=darkgray,
    thick,
    rounded corners=2pt,
    fill=white,
    minimum width=1.5cm,
    minimum height=0.65cm,
    align=center,
    font=\sffamily\small,
    right=0.7cm of stv
] (stvout) {
    StV
};

\node[
    draw=darkgray,
    thick,
    rounded corners=2pt,
    fill=white,
    minimum width=1.5cm,
    minimum height=0.65cm,
    align=center,
    font=\sffamily\small,
    below=2.5cm of penalty
] (score) {
    Score
};

\draw[
    ->,
    thick,
    draw=darkgray
] (val) -- (sim);


\draw[
    ->,
    thick,
    double,
    draw=darkgray
] (val) -- (lea);

\draw[
    ->,
    thick,
    draw=darkgray
] (correct) -- (misplaced);

\draw[
    ->,
    thick,
    draw=darkgray
] (misplaced) -- (pairing);

\draw[
    ->,
    thick,
    draw=darkgray
] (pairing) -- (redundant);

\draw[
    ->,
    thick,
    draw=darkgray
] (redundant) -- (stv);

\draw[
    ->,
    thick,
    double,
    draw=darkgray
] (pairing) -- (aqm);

\draw[
    ->,
    thick,
    double,
    draw=darkgray
] (redundant) -- (npaqm);

\draw[
    ->,
    thick,
    double,
    draw=darkgray
] (stv) -- (stvout);

\draw[
    ->,
    thick,
    double,
    draw=darkgray
] (redundant.south) -- ++(0,-0.2) -| (score.north);

\draw[
    ->,
    thick,
    double,
    draw=darkgray
] (lcs) -- ++(0,-1.9585) -| (score.north);

\draw[
    ->,
    thick,
    double,
    draw=darkgray
] (penalty) -- ++(0,-0.45) -| (score.north);

\end{tikzpicture}
\end{adjustbox}
\caption{Overview of the pipeline - evaluation metrics} 
\label{fig:plan_pipeline_metrics}
\end{figure}
We compare each $\pi_{\text{LLM}}$ to the chosen $\pi_{\text{GT}}$, to analyse the structure and semantics captured by the LLM. 
Thus, we assign each LLM-generated plan two scores: a similarity score using semantic similarity and LCS algorithms, and a score based on action quality labels. All computed metrics are depicted in Figure \ref{fig:plan_pipeline_metrics}. 

\paragraph{Planning-based Plan Validation.} 
%
All plans are validated using VAL \cite{howey2004val} and simulated based on VAL's grounded effects. From the initial state, we apply the effects of each action to the current state to create the next state. If an action violates domain constraints, the simulation ends.
We define a new metric, the Last Executable Action (LEA), which is the executable portion of a plan, i.e. $LEA(\pi) = e/|\pi|$, where $e$ is the number of actions that are executable before the first violation, or $e=|\pi|$ if there is no violation, such that $LEA=1$ means the plan is fully executable. We refer to the portion of plans that are fully executable from all plans in an experiment as Executability Rate (ER). The state simulation is later used for further analysis in Section \ref{sect:completion}. 
Back to our running example, VAL determines $\pi_\text{LLM}$ as invalid, with the first action of the plan being non-executable due to unmet preconditions (since \texttt{pick-up} action requires the block to be on the table), making LEA $=0$; the state simulation will thus include 1 state, which is the initial state.

\paragraph{Pairing actions.}
\begin{table}[ht!]
    \centering
    \caption{Action Quality Labels.}
    \label{tab:action_quality_labels}
    \setlength{\tabcolsep}{3pt}
    \renewcommand{\arraystretch}{1.2}
    \adjustbox{max width=0.96\linewidth}{
    \begin{tabular}{rllc}
        & Label & Description & Label score \\
        \midrule
        1 & \texttt{Correct} & The action is correct and in the same position as in $\pi_{\text{GT}}$ & 1 \\
        2 & \texttt{Misplaced} & Same action in different position to $\pi_{\text{GT}}$. & 0.5 \\
        3 & \texttt{SameAct} & Correct action name, parameters are not fully correct. & 0.25 \\
        4 & \texttt{DiffAct} & Wrong action name, some similarity in parameters. & 0.125 \\
        5 & \texttt{Redundant} &  No similarity to any unpaired actions in $\pi_{\text{GT}}$. & 0 \\
    \end{tabular}
    }
\end{table}
Each action in $\pi_{\text{LLM}}$ is paired with the most similar unpaired action in $\pi_{\text{GT}}$, if one exists. Pairing follows the label order in Table \ref{tab:action_quality_labels}, such that label (1) is paired first, and label (5) last. The table's label scores, used later in this section, are independent of the continuous similarity score used for pairing, as defined below. 

We perform one-to-one pairing, such that repeated actions in $\pi_{\text{LLM}}$ are not paired with the same action from $\pi_{\text{GT}}$. Pairing proceeds in exact-match passes for the first two labels, followed by the partial-match procedure below for the remaining actions. First, each action in $\pi_{\text{LLM}}$ is paired, in order, with an identical unpaired action in $\pi_{\text{GT}}$ that has the same position in the plan.
Remaining actions are then paired with an unpaired identical action in a different position. 
E.g., if $\pi_{\text{GT}}$ contains $a_1$ once, and $\pi_{\text{LLM}}$ contains $a_1$ twice, the first occurrence (in $\pi_\text{LLM}$ order) is eligible to pair with the single $\pi_\text{GT}$ instance as \texttt{Correct} or \texttt{Misplaced}; the second occurrence, finding no remaining unpaired exact action, is passed on to partial matching. This structure gives position-based precedence to actions in $\pi_{\text{LLM}}$.
We denote the ratio of exact to misplaced action and parameter matches as $w_1$, and set $w_1=0.5$. Thus, $sim(a,a', Correct) = 1$ while $sim(a,a', Misplaced) = 0.5 \cdot sim(a,a', Correct) = 0.5$.
Actions still unpaired are handled by labels (3) to (5), which require computing partial action similarity, as detailed in the next section.

\paragraph{Partial action similarity.}
Where actions in $\pi_{\text{LLM}}$ do not have an exact match in $\pi_{\text{GT}}$, we compute an action similarity score based on the semantic similarity of action names and parameters to find the best match. When object names are symbols, we employ parameter matching instead of semantic similarity.
Let $a$ and $a'$ be actions from $\pi_\text{LLM}$ and $\pi_\text{GT}$, respectively. $\alpha$ and $\alpha'$ denote action names, and $p=(p_1,\dots,p_n)$ and $p'=(p'_1,\dots,p'_m)$ represent the parameter tuples, for actions $a$ and $a'$, respectively. $W(a)=\sum_i \mathrm{meaningful}(p_i)$ counts semantically meaningful parameters, i.e. how many parameters are composed from words that exist as WordNet synsets, and $\sigma(s,s')$ measures semantic similarity between $s$ and $s'$. 
E, which computes overall semantic similarity for action parameters, is defined as:
$$
\text{E}(p,p') = \frac{1}{\max(|p|,|p'|)} \sum_{i=1}^{\min(|p|,|p'|)} {\sigma(p_i,p'_i)}
$$
and P, which computes overall parameter matchings, is defined as:
$$
\text{P}(p,p') = \frac{F + w_1\,M}{\max(|p|,|p'|)}
$$
where $F$ counts exact parameter matches, $M$ counts partial matches (i.e. wrong order).
The overall parameter alignment score C is defined as:
\[
\text{C}(p,p') =
\begin{cases}
1, & |p|=0 \text{ and } |p'|=0,\\
0, & |p|=0 \text{ or } |p'|=0,\\
\text{E}(p,p'), & W(a) > \frac{|p|}{2} \text{ and } W(a') > \frac{|p'|}{2},\\
\text{P}(p,p'), & \text{otherwise}.
\end{cases}
\]
Thus, the similarity for actions with partial matching is:
$$sim(a, a', label) = \bigl(\sigma(\alpha,\alpha') + C(p,p')\bigr)/2$$
$\sigma$ is computed using Wu-Palmer similarity \cite{meng2013semanticsimilarity} with WordNet, for which we split composed names into their components. We find that WordNet covers 99.99\% of action names in IPC domains; we detail coverage results and handling of edge cases in the Appendix.
Algorithms \ref{alg:get_action_pairs} and \ref{alg:best-match} detail the pairing process for labels (3) to (5), where each action in $\pi_{\text{LLM}}$ is paired to the most similar available action in $\pi_{\text{GT}}$ and the corresponding label is recorded in an action-quality-mapping (AQM); if $\pi_{\text{LLM}}$ is longer than $\pi_{\text{GT}}$, unpaired actions in $\pi_{\text{LLM}}$ are labelled as \texttt{Redundant}. 
In our running example, the accumulative action similarity score for $\pi_{\text{LLM}}$ is 1.87; score breakdown per action can be found in the Appendix.

\begin{algorithm}[tbp!]
\small
\caption{Pair remaining actions from $\pi_{\text{LLM}}$ and $\pi_{\text{GT}}$}
\label{alg:get_action_pairs}
\begin{algorithmic}[1]
\Require unmatched candidate actions $A$ from $\pi_\text{LLM}$, unmatched ground-truth actions $A'$ from $\pi_\text{GT}$, pair list $P$, and  $AQM$
\Ensure final $P$ and $AQM$

\For{each $\ell \in (\textsc{SameAct}, \textsc{DiffAct})$} \Comment{label order (3), (4)}
    \For{each $a \in A$, in order, with index in $\pi_\text{LLM}$ $i = \mathrm{idx}(a)$}
        \If{$AQM(i)$ is defined \textbf{or} $A' = \emptyset$ \textbf{or} $|P| \ge |\pi_\text{GT}|$}
            \State \textbf{continue} \Comment{already paired, or no GT actions left}
        \EndIf
        \State $(a^*, \mathit{similarity}) \gets \textsc{BestMatch}(a, i, A', \ell)$ \Comment{Algorithm~\ref{alg:best-match}}
        \If{$a^* \neq \bot$}
            \State remove $a^*$ from $A'$; $P.\text{append}((a, a^*, \mathit{similarity}))$; 
            \State $AQM(i) \gets \ell$
        \EndIf
    \EndFor
\EndFor
\State mark every $a \in A$ with undefined label as $\textsc{Redundant}$ in $AQM$
\State \Return $(P, AQM)$
\end{algorithmic}
\end{algorithm}

\begin{algorithm}[tbp!]
\small
\caption{$\textsc{BestMatch}(a, i, A', \ell)$}
\label{alg:best-match}
\begin{algorithmic}[1]
\State $\mathit{similarity} \gets 0$, \quad $a^* \gets \bot$
\For{each $a' \in A'$}
    \If{$\bigl(\alpha = \alpha'\bigr) \neq (\ell = \textsc{SameAct})$}
        \State \textbf{continue} \Comment{action does not belong to this label}
    \EndIf
    \State $s \gets sim(a, a', \ell)$
    \State $s \gets s$ \textbf{if} $\mathrm{idx}_\text{GT}(a') = i$ \textbf{else} $s \cdot w_1$ \Comment{positional discount}
    \If{$s > \mathit{similarity}$} \State $\mathit{similarity} \gets s$, \quad $a^* \gets a'$ \EndIf
\EndFor
\State \Return $(a^*, \mathit{similarity})$
\end{algorithmic}
\end{algorithm}

\paragraph{LCS algorithms.} \label{sect:lcs}
We further assess plan similarity using two types of LCS algorithms \cite{lcs_algorithms}. 
The longest common contiguous subsequence between $\pi_{\text{LLM}}$ and $\pi_{\text{GT}}$ (denoted as $LCCS(\pi, \pi')$), and the longest common ordered subsequence of actions, even when those are not contiguous (denoted as $LCOS(\pi, \pi')$). 
We compute an LCS score, which is defined as:
$$
LCS(\pi, \pi_\text{GT}) = \Bigl(w_2 \cdot |LCCS(\pi, \pi_\text{GT})| + (1-w_2) \cdot |LCOS(\pi, \pi_\text{GT})|\Bigr)
$$
We set $w_2=0.6\dot{6}$, as a contiguous subsequence is harder to achieve and is a stronger signal in terms of reasoning capabilities.
For our example $\pi_{\text{LLM}}$, $|LCCS(\pi_{\text{LLM}}, \pi_\text{GT})| = 1$, and $|LCOS(\pi_{\text{LLM}}, \pi_\text{GT})| = 2$, so $LCS(\pi_{\text{LLM}}, \pi_\text{GT}) = 1.3\dot{3}$.

\paragraph{AQM and NP-AQM} \label{sect:aqm}
As an additional action quality measure, we assign each label a rank based on its quality -- the \texttt{Redundant} label gets a score of 0, as these actions do not contribute to the plan; correct actions are given the highest score, which is the max score an action can get through similarity evaluation, and other labels are given exponentially smaller scores. The scores per label are specified in Table \ref{tab:action_quality_labels}. 
As in Algorithm \ref{alg:get_action_pairs}, we track the label of each action in $\pi_{\text{LLM}}$ with a mapping (AQM). We compute an AQM score as a mean of ranks for the labels associated with all actions in the plan. 
In our example, $\pi_{\text{LLM}}$ will receive the following AQM:
\texttt{\small \{1: SameAct, 2: Correct, 3: Misplaced, 4: DiffAct\}}, and an AQM score of $0.47$. 
To investigate position-based bias in action quality throughout plan generation, we create a complementary non-positional AQM (NP-AQM), which disregards action order and one-to-one pairing of GT actions. \texttt{\small Misplaced} actions are relabelled as \texttt{\small Correct}, and each action labelled \texttt{\small Redundant} in the AQM is relabelled by comparing it to all actions in $\pi_{\text{GT}}$, regardless of whether they were already paired: it becomes \texttt{\small Correct} if it is identical to a GT action, \texttt{\small SameAct} if it shares an action name with one, \texttt{\small DiffAct} if it otherwise resembles one, and remains \texttt{\small Redundant} only if it resembles none. This distinction enables the assessment of both strict action-order compliance and general action correctness. NP-AQM is used only for analysis and does not contribute to a score metric.
Accordingly, the resulting NP-AQM is:
\texttt{\small \{1: SameAct, 2: Correct, 3: Correct, 4: DiffAct\}}.

\paragraph{Length-based factor.} \label{sect:length}
We penalise plans whose length deviates from a reference plan, with shorter plans penalised more heavily, as they are more likely to be missing required actions. For a plan $\pi$, reference plan $\pi_\text{ref}$, and threshold plan $\pi_\text{thr}$, we define the length-deviation penalty as
$$
\delta(\pi, \pi_\text{ref}, \pi_\text{thr}) = \frac{\text{abs}\bigl(|\pi|-|\pi_\text{ref}|\bigr)}{\kappa}, \qquad
\kappa = \begin{cases} 1, & |\pi| < |\pi_\text{thr}|,\\ 2, & \text{otherwise} \end{cases}
$$
with $\delta = 0$ if $\pi$ and $\pi_\text{ref}$ are both empty. $\delta$ is used in two forms: raw, to rank GT plan candidates against $\pi_\text{LLM}$ (Section~\ref{sect:choosing-gt}), and normalised, as part of the length factor $LF$ below.
For choosing a GT plan, we define LP:
$$
LP = \frac{2 \cdot \delta(\pi_\text{LLM}, \pi_{\text{GT}_n}, \pi_{\text{GT}_n})}{|\pi_\text{LLM}|}, \qquad \pi_\text{LLM} \neq \emptyset 
$$

For valid plans, we only penalise sub-optimality, since $\pi^*$ is generated using FD with A* and the LM-Cut heuristic to find an optimal plan. For invalid plans, we use $\pi^*$ as the threshold, since $\pi^*$ is the optimal (and therefore shortest valid) plan, so a plan shorter than $\pi^*$ is necessarily missing actions. If exactly one of $\pi$ and $\pi_\text{GT}$ is empty, $LF=0$. Otherwise,
\begin{align*}
LF_\text{valid}(\pi)   &= \min\Bigl(1,\ \frac{|\pi^*|}{|\pi|}\Bigr), \qquad \pi \neq \emptyset,\\[0pt]
LF_\text{invalid}(\pi) &= 1 - \frac{\delta(\pi, \pi_\text{GT}, \pi^*)}{\max(|\pi|,|\pi_\text{GT}|)}.
\end{align*}
$LF(\pi)$ equals $LF_\text{valid}(\pi)$ if $\pi$ is valid, and $LF_\text{invalid}(\pi)$ otherwise. In our running example, $\pi_\text{LLM}$ is the same length as $\pi_\text{GT}$, so no penalty is applied. 

\paragraph{Similarity score} \label{sect:similarity-score}
To initially choose $\pi_\text{GT}$, we use a simplified heuristic of the plan evaluation module:
\begin{align*}
S_n = LCS(\pi_\text{LLM},\pi_{\text{GT}_n})  - & LP(\pi_\text{LLM}, \pi_{\text{GT}_n}), \qquad \pi_\text{LLM} \neq \emptyset,\\[0pt]
S_n = -|\pi_{\text{GT}_n}|, & \qquad \pi_\text{LLM} = \emptyset
\end{align*}
Because the length penalty is divided by $|\pi_\text{LLM}|$, it acts mainly as a tie-breaker between candidates with similar LCS. An empty $\pi_\text{LLM}$ will have the shortest GT plan as its $\pi_\text{GT}$.
In the evaluation pipeline, the similarity score for LLM-generated plans is computed as:
\begin{align*}
\text{S}_\text{valid}(\pi) &= LF_\text{valid}(\pi), \\[0pt]
\text{S}_\text{invalid}(\pi) &= \frac{LF_\text{invalid}(\pi)}{2|\pi|}\left[ \bigl(\sum_{i=1}^{|\pi|} sim(a_i,a_i^*,label_i) + LCS(\pi, \pi_\text{GT}) \right].
\end{align*}
where $a_i^*$ is the GT action paired with $a_i$, and all scores are in range $[0,1]$.
For our example plan, $S_\text{invalid}(\pi_\text{LLM})=0.4$. 

\subsection{Plan Potential} \label{nlp-manipulation} 
To better understand LLM's plan generation behaviour, we define plan potential as the closest a plan can get to $\pi_{\text{GT}}$ using simple NLP manipulation, i.e. $S(\pi_\text{NLP})$. The reasoning is that if LLMs are pattern-matching when generating a plan, simple transformations can likely recover a valid plan, assuming the LLM has generated one that solves a similar problem. \review{explanation makes sense? wording appropriate?}
When $\pi_{\text{LLM}}$ is valid, we do not need to manipulate it into a valid plan. For invalid plans, we compute a holistic score which captures various aspects of plan quality, as the average of the previous plan evaluation score, $S(\pi_\text{LLM})$, and plan potential. 

\paragraph{NLP Manipulation.}
We calculate plan potential by methodically applying two basic NLP transformations, grounded in mathematical operations from existing approaches for refining sequences \cite{jiraskova2008state,panda2021shuffled}: (1) left circular-shifts, which use modulo arithmetic to move each action by $n$ steps; and (2) consistent parameter remapping to remap plan parameters consistently across all actions, using the Hungarian method \cite{kuhn1955hungarian}, which solves the assignment problem in polynomial time, to find the best alignment to $\pi_{\text{GT}}$. We define this alignment as maximising matches between parameters of the two plans. In typed domains, mappings are restricted to parameters of the same type, based on the problem definition. Otherwise, all objects are considered of the same type. 
To avoid a combinatorial explosion, plan variants are ranked with a preliminary score that counts, for each position, action-name matches and exact action matches with $\pi_{\text{GT}}$. Only the top-$c$ candidates are assessed with the plan evaluation approach. Valid plans are prioritised such that if one is found, it is chosen directly. Otherwise, the invalid candidate with the highest score becomes $\pi_{\text{NLP}}$ (or $\pi_{\text{LLM}}$ if no candidate scores higher). $c$ is chosen using a hyperparameter search. Further details on the implementation of the Hungarian method and the search can be found in the Appendix.
In our example, $\pi_{\text{NLP}}$ has 2 circular shifts, and mapping \texttt{\small \{a: c, b: a, c: b\}} which results in:
\noindent \texttt{[\small (stack a b), (stack c a), (pick-up b), (stack b c)]}. This plan receives a higher score $S(\pi_\text{NLP}) = 0.77$. 
The final score for $\pi_{\text{LLM}}$ is the mean of the evaluation scores of $\pi_{\text{LLM}}$ and $\pi_{\text{NLP}}$, i.e. the potential of $\pi_{\text{LLM}}$, which in our example is $Score(\pi_\text{LLM}) = 0.59$. 

\subsection{Plan Refinement}
We present a hybrid method for plan refinement to extract additional information about the quality of LLM-generated plans.
We utilise planning tools and a planning-based analysis, together with the NLP evaluation, to refine $\pi_{\text{LLM}}$ and $\pi_{\text{NLP}}$.

\paragraph{Best LCS} \label{best-lcs} 
If $LCCS(\pi_{\text{LLM}}, \pi_\text{GT})$ is a valid plan, or reaches the goal state, we choose it as the best LCS, and truncate it if the goal is not the last state. Otherwise, we perform the same for $LCOS(\pi_{\text{LLM}}, \pi_\text{GT})$. If neither reaches the goal, the LCOS is selected, as it is at least as long as the LCCS and therefore retains at least as many correct actions.
In our example, for $\pi_{\text{LLM}}$, $LCCS(\pi_{\text{LLM}}, \pi_\text{GT})$ is invalid, and $LCOS(\pi_{\text{LLM}}, \pi_\text{GT})$ consists of two actions: \texttt{[\small (stack c a), (stack b c)]}. Nonetheless, for $\pi_{\text{NLP}}$, the best LCS is valid, and is equal to $\pi^*$: \texttt{\small [(pick-up b), (stack b c)]}. 
These plans are denoted as $\pi_{LCS}$ and $\pi_\text{NLP-LCS}$ respectively. 

\paragraph{Steps to validity.}
To quantify the effort required to make plans valid, we perform stepwise modifications to transform a plan into $\pi_{\text{GT}}$. 
Edit distance is a common metric for calculating similarity between two sequences, providing further indication of how far LLM plans are from those of a symbolic planner, by considering them as token output. 
We define the edit distance between two plans with move operations, where each operation (reorder, repair, add, and remove) is assigned equal cost. 
To find the set of changes required, we use AQM's labels to guide the process.
The number of operations is the edit distance and provides an additional metric of plan quality, as an upper bound for how far a plan is from validity. We refer to the metric as Steps to Validity (StV); its computation algorithm is presented in the Appendix.
Our example $\pi_{\text{LLM}}$ requires 3 steps, while $\pi_{\text{NLP}}$ requires only 1.

\paragraph{Planning-based plan completion.} \label{sect:completion} 
We utilise a symbolic planner to provide further insights into the use of generated plans in advancing towards the goal.
We start by using the state simulations for $\pi_{\text{GT}}$ and $\pi_\text{LLM}$ to find the point (if it exists) at which the two plans' states diverge, which we consider as the correct part of the plan, as it undoubtedly advances towards the goal. We achieve this by identifying the longest prefix of $\pi_{\text{GT}}$'s state simulation that appears, in order, in $\pi_{\text{LLM}}$'s state simulation. For the example $\pi_{\text{LLM}}$, this is the initial state, as no other state exists after. We refer to this sub-plan in $\pi_{\text{LLM}}$ as $\pi_{\text{corr}}$. 
The final step of the process involves replanning, for which we reuse Fast-Downward with the same search settings. The last state of $\pi_{\text{corr}}$ is set as the initial state in a new PDDL problem with the original goal, and the resulting sub-plan completes $\pi_\text{corr}$ towards validity. We refer to this plan as the complementary plan and denote it as $\pi_{\text{comp}}$. The plan comprising these two sub-plans is set as $\pi_\text{VAL}$. $\pi_{\text{comp}}$ in our example is equal to $\pi^*$, as no state has been advanced with the actions of $\pi_{\text{LLM}}$. We use $\pi_{\text{comp}}$ as an additional indication of how far $\pi_{\text{LLM}}$ is from a planner-generated plan. 
\section{Evaluation}
We evaluate plans generated by five LLM families across four prompt types, four domain with five problem suites, and a hundred problem instances each.
\subsection{Experimental Setup}
HuggingFace\footnote{\url{https://huggingface.co/}} is used to integrate open-source models into the pipeline. The models chosen are those that generate any plans in preliminary zero-shot experiments (Qwen and Gemma). 
For proprietary models, we incorporate API calls to Anthropic and Google Cloud in addition to the existing OpenAI calls. We generate plans using frontier Anthropic, Google Cloud, and OpenAI models. 
All generations were conducted between June and September 2026. For further analysis of model behaviour, we also include older LLMs, for which generation was conducted between March and July 2025; when those are included, it is clearly stated. A full list of all used models can be found in the Appendix. 
We maintain PlanBench's default value of top-p=1 and, when supported, a temperature of 0; otherwise, if temperature can be set, we use a minimal value of 0.1, or, in recent models which do not support temperature settings, we maintain the default configurations. We set a timeout for plan generation of 30 minutes per instance, and limit thinking and output tokens for open models with the CoT prompt type to 4096 and 1024, respectively.
We use Ubuntu 22.04 computers with 40 to 80 GB VRAM to run the plan generation pipeline on open-source models, and an Ubuntu 24.04 computer with 64GB of RAM and 28 cores for API interaction, and to run our pipeline.
In cases where plan generation was unsuccessful, and evaluation is therefore impossible, these are discarded. Across all experiments (16000 expected plans), 5,479 produced no output or have timed-out, or produced output that could not be parsed into any plan. We discard the failed plans, leaving a total of 10,521 generated plans to evaluate. 

\paragraph{Prompt types and domains} \label{sect:tasks-domains}
We generate plans for four prompt types: NL zero-shot, NL one-shot, NL state-tracking (CoT), and PDDL one-shot, as they are set in PlanBench. 
We use the provided Blocksworld (BW) domain \cite{gupta1992complexity} and Logistics domain \cite{mcdermott20001998}, and incorporate two additional IPC domains: Elevator \cite{koehler2000elevator} and Satellite \cite{long20033rd}. For BW, we use both problem suites from PlanBench: an easier version which uses only 3 blocks (Blocksworld 3), and a harder version with more blocks (Blocksworld). 
The results in the main paper refer to the mean results for all domains, unless stated otherwise. Results for each domain and further general analysis are included in the Appendix.

\paragraph{Tools for plan evaluation and refinement.}
Semantic similarity is computed using NLTK's implementation of Wu-Palmer similarity measure with WordNet \cite{meng2013semanticsimilarity}.
For state simulation, we employ Tarski\footnote{\url{https://github.com/aig-upf/tarski}}, a PDDL parser, and VAL, which is also used to compute plan validity and executability. 
In the completion stage, we reuse FD to replan, ensuring consistency with the generation of GT plans whilst ensuring optimality of $\pi_\text{comp}$. This would be equivalent to running SymK with $k=1$. 

\begin{figure*}[!htb]
    \centering
    \begin{subfigure}[t]{0.48\textwidth}
        \centering
        \includegraphics[height=3.8cm]{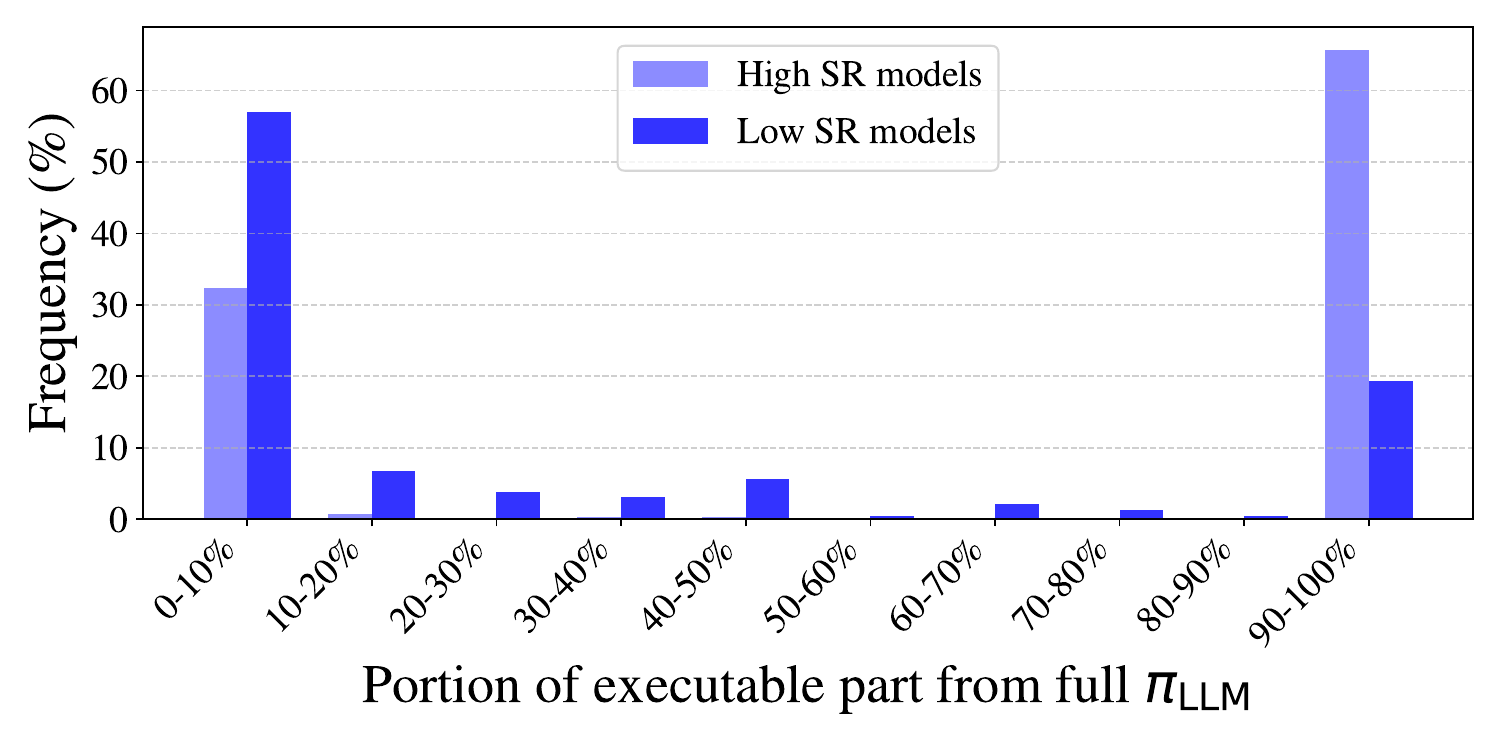}
        \caption{Distribution of LEA across the length of $\pi_\text{LLM}$ across domains} 
        \label{fig:last_exec_act_as_percentage}
    \end{subfigure}
    ~
    \begin{subfigure}[t]{0.48\textwidth}
        \centering
        
        \includegraphics[height=3.8cm]{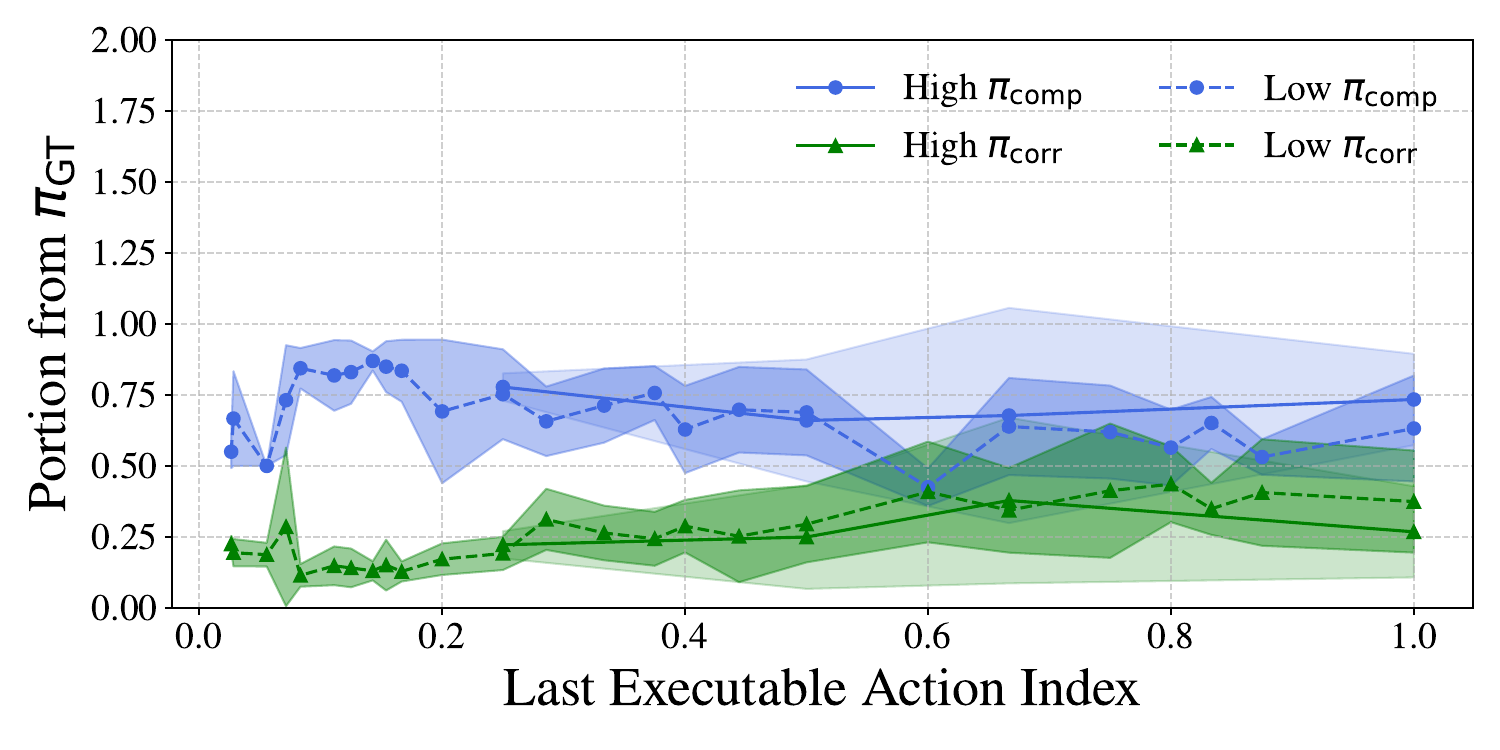}
        \caption{LLM VS planner contribution to $\pi_\text{VAL}$ across domains}
        
        \label{fig:comp_plan_vs_gt_plan}
    \end{subfigure}
    \label{fig:SR_per_GT_length}
    \caption{Quality Analysis through plan refinement}
\end{figure*}
\subsection{Results} 
A table with formal definitions of all metrics presented in the paper is included in the Appendix, along with results with $\pi_\text{GT} = \pi^*$ (i.e. $k=1$ for SymK), which are largely comparable. 

\paragraph{Plan quality.} 
Figure \ref{fig:last_exec_act_as_percentage} shows the distribution of LEA for $|\pi_\text{LLM}|$ across experiments. 
Across all plans, for almost half of the plans (48.13\%), $\text{LEA} < 0.5$. The first action violates domain constraints ($\text{LEA}=0$) in 38.92\% of all plans, and 58.3\% of invalid plans. 
As explained in Section~\ref{sect:completion}, the final stage includes replanning, to complete $\pi_{\text{corr}}$ with $\pi_{\text{comp}}$, resulting in $\pi_\text{VAL}$. 
Figure \ref{fig:comp_plan_vs_gt_plan} further inspects these sub-plans for cases where $\pi_{\text{comp}}$ is generated, and the LLM has advanced at least one state towards the goal, i.e. $|\pi_{\text{corr}}| > 0$. 
We find that for 60.46\% of plans, $|\pi_{\text{comp}}| = |\pi^*|$ and $|\pi_\text{corr}| = 0$, and even for plans with a high LEA $\frac{|\pi_{\text{corr}}|}{|\pi_\text{VAL}|} < 0.5$ in most cases. These fidings make it clear that $|\pi_{\text{corr}}|$ does not significantly grow with the LEA of $\pi_\text{LLM}$.

\paragraph{Plan refinement.}
We inspect improvement in validity from $\pi_\text{LLM}$ to $\pi_\text{NLP-LCS}$, and compare patterns of current LLMs with LLMs from 2025. For the comparison, we look at the BW3 and Logistics domain and problem suites. 
Our results show that the NLP-manipulation and refinement pipeline is more effective for shorter $\pi^*$s, with complex problems remaining a challenge and thus further away from becoming valid. With recent LLMs, we find that SR across all 5 problem suites improves from 0.61 to 0.71 for $|\pi^*|=2$, a 16\% improvement. This drops to 4.3\% for $|\pi^*|=4$, and 0\% improvement for $|\pi^*|=6$ and $|\pi^*|=8$. For BW3 and Logistics only, these become 11.1\%, 6.1\%, 0\%, and 2.4\%, respectively. 
This improvement is significantly lower than comparable results for 2025 LLMs, which have showcased improvement of 27.6\%, 17.7\%, 13.5\%, and 2.5\%, respectively on BW3 and Logistics.
Improvement results by model can be found in the Appendix. 

\paragraph{Effectiveness of metrics.} 
\begin{table*}[tb!]
\caption{Comparison of core metrics by domain}
\label{tab:core_metrics_by_domain}
\resizebox{\textwidth}{!}{
\begin{tabular}[0.9\textwidth]{lccccccccccc}
\toprule
Domain & $\pi_{\text{LLM}}$ SR $\uparrow$ & $\pi_{\text{LLM}}$ ER $\uparrow$ & Score $\uparrow$ & AQM $\uparrow$ & StV $\downarrow$ & LEA $\uparrow$ & $|\pi_{\text{corr}}|$ $\uparrow$ & $\pi_{\text{NLP-LCS}}$ SR $\uparrow$ & $\pi_{\text{NLP-LCS}}$ StV $\downarrow$ & $\pi_{\text{NLP-LCS}}$ LEA & $\pi_{\text{VAL}}$ SR $\uparrow$ \\
\midrule
Blocksworld & 0.3 & 0.34 & 0.43 & 0.49 & 9.52 & 0.37 & 2.33 & 0.32 & 5.41 & 0.46 & \textbf{1.0} \\
Blocksworld 3 & \textbf{0.56} & 0.7 & \textbf{0.73} & \textbf{0.83} & \textbf{2.8} & \textbf{0.77} & \textbf{3.07} & \textbf{0.58} & \textbf{1.91} & \textbf{0.81} & \textbf{1.0} \\
Elevator & 0.47 & \textbf{0.71} & 0.6 & 0.53 & 5.61 & 0.75 & 1.02 & 0.47 & 5.02 & 0.8 & \textbf{1.0} \\
Logistics & 0.12 & 0.15 & 0.33 & 0.32 & 14.54 & 0.17 & 0.43 & 0.13 & 12.71 & 0.25 & \textbf{1.0} \\
Satellite & 0.22 & 0.56 & 0.39 & 0.38 & 12.9 & 0.61 & 0.58 & 0.22 & 11.58 & 0.55 & \textbf{1.0} \\
\bottomrule
\end{tabular}
}
\end{table*}

\begin{table*}[tb!]
\caption{Comparison of core metrics by model - Satellite}
\label{tab:core_metrics_by_model}
\resizebox{\textwidth}{!}{
\begin{tabular}[0.9\textwidth]{lccccccccccc}
\toprule
Model & $\pi_{\text{LLM}}$ SR $\uparrow$ & $\pi_{\text{LLM}}$ ER $\uparrow$ & Score $\uparrow$ & AQM $\uparrow$ & StV $\downarrow$ & LEA $\uparrow$ & $|\pi_{\text{corr}}|$ $\uparrow$ & $\pi_{\text{NLP-LCS}}$ SR $\uparrow$ & $\pi_{\text{NLP-LCS}}$ StV $\downarrow$ & $\pi_{\text{NLP-LCS}}$ LEA & $\pi_{\text{VAL}}$ SR $\uparrow$ \\
\midrule
Claude Opus 4.8 & \textbf{0.39} & 0.74 & 0.53 & 0.33 & 13.29 & 0.79 & 0.52 & \textbf{0.39} & 12.4 & 0.63 & \textbf{1.0} \\
GPT 6 Sol & \textbf{0.39} & \textbf{0.97} & \textbf{0.6} & 0.44 & \textbf{8.35} & \textbf{0.97} & \textbf{1.38} & \textbf{0.39} & \textbf{8.16} & \textbf{0.71} & \textbf{1.0} \\
Gemini 3.1 Pro & \textbf{0.39} & 0.54 & 0.53 & 0.43 & 9.03 & 0.55 & 0.58 & \textbf{0.39} & 9.16 & 0.5 & \textbf{1.0} \\
Gemini 3.8 Flash & 0.0 & 0.59 & 0.06 & \textbf{0.46} & 14.16 & 0.59 & 0.12 & 0.0 & 14.15 & 0.67 & \textbf{1.0} \\
\hline \hline
Gemma 4 12B & 0.02 & 0.25 & 0.26 & 0.31 & 16.56 & 0.37 & 0.24 & 0.02 & 13.65 & 0.39 & \textbf{1.0} \\
Gemma 4 26B & 0.0 & 0.17 & 0.12 & 0.23 & 20.92 & 0.31 & 0.13 & 0.0 & 13.66 & 0.51 & \textbf{1.0} \\
Qwen3 14B & 0.0 & 0.16 & 0.24 & 0.36 & 15.04 & 0.25 & 0.15 & 0.0 & 13.8 & 0.34 & \textbf{1.0} \\
Qwen3 8B & 0.01 & 0.21 & 0.28 & 0.32 & 15.72 & 0.34 & 0.45 & 0.01 & 13.83 & 0.4 & \textbf{1.0} \\
\bottomrule
\end{tabular}
}
\end{table*}

\begin{table*}[tb!]
\caption{Comparison of core metrics by prompt type - Satellite}
\label{tab:core_metrics_by_task}
\resizebox{\textwidth}{!}{
\begin{tabular}[0.9\textwidth]{lccccccccccc}
\toprule
Prompt Type & $\pi_{\text{LLM}}$ SR $\uparrow$ & $\pi_{\text{LLM}}$ ER $\uparrow$ & Score $\uparrow$ & AQM $\uparrow$ & StV $\downarrow$ & LEA $\uparrow$ & $|\pi_{\text{corr}}|$ $\uparrow$ & $\pi_{\text{NLP-LCS}}$ SR $\uparrow$ & $\pi_{\text{NLP-LCS}}$ StV $\downarrow$ & $\pi_{\text{NLP-LCS}}$ LEA & $\pi_{\text{VAL}}$ SR $\uparrow$ \\
\midrule
One-Shot & 0.1 & 0.44 & 0.33 & 0.39 & 14.3 & 0.53 & 0.6 & 0.1 & 12.71 & 0.49 & \textbf{1.0} \\
PDDL & \textbf{0.57} & 0.65 & \textbf{0.63} & 0.33 & \textbf{9.78} & 0.68 & 0.3 & \textbf{0.57} & \textbf{8.29} & \textbf{0.77} & \textbf{1.0} \\
CoT & 0.17 & \textbf{0.96} & 0.42 & \textbf{0.41} & 13.52 & \textbf{0.96} & \textbf{1.13} & 0.17 & 12.99 & 0.57 & \textbf{1.0} \\
Zero-Shot & 0.03 & 0.51 & 0.19 & 0.39 & 13.98 & 0.54 & 0.65 & 0.03 & 13.1 & 0.4 & \textbf{1.0} \\
\bottomrule
\end{tabular}
}
\end{table*}

Table \ref{tab:core_metrics_by_domain} compares performance across domains using the metrics introduced in the pipeline, and SR and ER as baselines. Metric values (excluding StV and $|\pi_\text{corr}|$) are in range $[0,1]$.
When evaluating plan quality in a generalised manner, that is, grouped by domain with all tasks and models aggregated, plan quality is significantly distinct across domains, and SR is mostly consistent with other metrics.
However, when inspecting less generic environments, for example, by evaluating one domain at a time, we can see that these patterns are less obvious, and that ER and SR are both less consistent with other metrics, and with the quality of refined plans. Tables  \ref{tab:core_metrics_by_model} and \ref{tab:core_metrics_by_task} present evaluation results for the Satellite domain, which has shown no improvement in plan validity using the refinement pipeline. 
We can see that there is high variance in LLMs ability to generate valid plans (SR) and executable plans (ER), across both models and prompt types. 
For example, Table~\ref{tab:core_metrics_by_model} shows that although Claude Opus 4.8 and GPT 6 Sol are comparable by SR, ER and Score are much higher for the GPT model. Here, ER and Score are more consistent with StV and LEA values, which shows that Score, an NLP-based metric, is able to capture diverse aspects of plan generation. 
In Table \ref{tab:core_metrics_by_task}, the prompt type with the highest initial SR, one-shot PDDL, does indeed have the best performing $\pi_\text{NLP-LCS}$ plans across metrics. However, CoT achieves a much better ER, which makes ER unaligned here with most metrics. This suggests that neither SR nor ER on their own are satisfactory to investigate the intricacies and different aspects of model behaviour in plan generation.
When computing per-plan mean Pearson correlation of primary metrics (SR, ER, and Score) with all other metrics, we find that Score achieves a comparable correlation to SR and ER. When comparing by groups, we find the Score's correlation is stronger for 4/5 domains, 7/8 models, and 4/4 prompt types. When considering the overall ungrouped correlation, Score achieves the highest correlation (0.628), SR comes second (0.598), and ER last (0.583). A bootstrap analysis of difference in correlation shows that the 95\% CI for Score-SR is [0.026, 0.035], and Score-ER is [0.036, 0.055], both excluding 0. Tables with the correlation details can be found in the Appendix. 
\section{Discussion} 
We show that using our pipeline, we can provide a more granular plan evaluation, that strongly correlates to existing evaluation methods, and is able to reveal generation patterns across LLMs, which are not always consistent with SR and ER.
We reiterate that we aim to evaluate and investigate plan quality and LLM-behaviour, and that the evaluation is not intended to be used as an LLM-based planning pipeline; hence, the pipeline relies on ground-truth plans produced by symbolic planners, and the LLM is no longer independent. 
Across all investigated LLMs, most failures in generated plans cannot be fixed through NLP-based refinement, implying that LLMs do not consistently generate nearly-correct solutions that only require post-processing, and plan quality is highly varied.
It is also evident that models considered of better quality generally have higher SR. However, it is noteworthy that this does not imply the invalid plans they generate are also consistently better, as evidenced by the results of other metrics, for example, when comparing Gemma4 12B and Qwen3 8B on the Satellite domain.
This suggests that a more granular approach to evaluating plan quality can lead to a better choice of model for a specific task. It is also evident from the results that our Score metric highly correlates to other NLP- and planning-based evaluation metrics, making it a useful independent metric. 
The score, or a suite of metrics, could therefore be used as a better signal for training LLMs, or for a holistic evaluation and analysis of models' behaviour and weaknesses, with the ability to focus on one plan at a time, without relying on a binary classification. 

The analysis reveals that in the vast majority of non-executable plans, no actions in the plan are executable at all, suggesting that in these cases, the LLM does not successfully reason over the search space, but rather generates superficially plausible actions. 
In older LLMs, applying simple transformations, such as parameter swaps and reordering of actions, often leads to improvement across metrics.
This may imply that the model does not construct plans from logical derivation, but replicates or adapts examples it has seen during training or in the prompt. This pattern has largely changed with most recent models, which may suggest a general change in LLMs' behaviour in planning. 
Lastly, we see that even in highly executable plans, LLMs still struggle with advancing towards the goal, as $\pi_\text{comp}$ is still a significant portion of $\pi_{VAL}$. The high number of cases where $|\pi_{\text{comp}}| = |\pi^*|$ raises doubt on whether LLM plans can be used as a useful starting point for a valid plan.
\section{Conclusion}
We have introduced a new evaluation pipeline for investigating the quality of LLM-generated plans and patterns in the plan generation task, which demonstrates that SR is insufficient as an indicator of plan quality. 
We show that conducting an analysis of the output using both NLP- and planning-based methods can provide further insights into the cases where LLMs struggle.
The high variability in improvement across experiments, which does not strictly align with the initial SR and ER of the model, suggests that there is more to plan quality than can be captured with binary classification. 
Our results show that our metrics could be leveraged for further investigation of model behaviour, or to provide reward signals or explanations for improving LLM planning capabilities.
Models' capabilities improve over time; however, their planning capabilities are still not as reliable as symbolic planners, which require a state-tracking mechanism that LLMs are not built for; a classical planner is thus necessary. 
LLM plan quality may be influenced by the models' architectures, and the relation between the two could be further explored.
Future work could further explore LLMs' temporal planning ability. 


\begin{acks}
This work has been supported by UK Research and Innovation (EP/S023356/1), in the UKRI Centre for Doctoral Training in Safe and Trusted Artificial Intelligence (www.safeandtrustedai.org).
This work has also been supported by King's College London (2025), King's Computational Research, Engineering and Technology Environment (CREATE). Retrieved March 22, 2025, from \url{https://doi.org/10.18742/rnvf-m076}.
\end{acks}



\bibliographystyle{ACM-Reference-Format} 
\bibliography{main}

\appendix
\onecolumn
\renewcommand{\thefigure}{A.\arabic{figure}}
\renewcommand{\thetable}{A.\arabic{table}}
\clearpage
\begin{center}
{\huge\bfseries Supplementary Material \par}
\end{center}
\vspace{1em}

\setcounter{figure}{0}
\setcounter{table}{0}
\section*{Appendix A: Full evaluation results}
\subsection*{Full results.} Tables \ref{tab:metric_comparison_by_task_all} to \ref{tab:core_metrics_by_model_elevator} present the aggregated results for model and prompt type by domain. Tables \ref{tab:appendix_metrics_per_experiment_by_domain} and \ref{tab:appendix_metrics_per_experiment_by_domain_2} provide the full results for each configuration (domains x model × prompt type), covering all evaluation metrics.

\subsection*{Optimal plan as GT}
We report the results of running our pipeline with an optimal and valid plan as the ground truth, which is done by skipping the choosing GT plan module (and equivalent to using te module with k=1). Tables \ref{tab:core_metrics_by_domain_opt}, \ref{tab:core_metrics_by_model_opt}, and  \ref{tab:core_metrics_by_task_opt} provided the full results aggregated by domain, model, and prompt type, respoectively.

\begin{table}[b]
\caption{Comparison of core metrics by domain - optimal plan as GT}
\label{tab:core_metrics_by_domain_opt}
\resizebox{\textwidth}{!}{
\begin{tabular}[0.9\textwidth]{lcccccccccccc}
\toprule
Domain & $\pi_{\text{LLM}}$ SR $\uparrow$ & $\pi_{\text{LLM}}$ ER $\uparrow$ & Score $\uparrow$ & AQM $\uparrow$ & StV $\downarrow$ & LEA $\uparrow$ & $|\pi_{\text{corr}}|$ $\uparrow$ & $\pi_{\text{NLP-LCS}}$ SR $\uparrow$ & $\pi_{\text{NLP-LCS}}$ StV $\downarrow$ & $\pi_{\text{NLP-LCS}}$ LEA & $\pi_{\text{VAL}}$ SR $\uparrow$ \\
\midrule
Blocksworld & 0.3 & 0.34 & \textbf{0.73} & 0.79 & 6.2 & 0.37 & 1.71 & 0.31 & 3.74 & 0.4 & \textbf{1.0} \\
Blocksworld 3 & \textbf{0.56} & 0.7 & 0.72 & \textbf{0.82} & \textbf{2.8} & \textbf{0.77} & \textbf{3.04} & \textbf{0.58} & \textbf{1.85} & 0.8 & \textbf{1.0} \\
Elevator & 0.47 & \textbf{0.71} & 0.6 & 0.55 & 5.3 & 0.75 & 1.39 & 0.47 & 4.35 & \textbf{0.81} & \textbf{1.0} \\
Logistics & 0.12 & 0.16 & 0.33 & 0.31 & 14.18 & 0.17 & 0.38 & 0.13 & 12.47 & 0.24 & \textbf{1.0} \\
Satellite & 0.22 & 0.56 & 0.39 & 0.38 & 12.87 & 0.61 & 0.81 & 0.22 & 11.53 & 0.54 & \textbf{1.0} \\
\bottomrule
\end{tabular}
}
\end{table}

\begin{table}[b]
\caption{Comparison of core metrics by model (all domains) - optimal plan as GT}
\label{tab:core_metrics_by_model_opt}
\resizebox{\textwidth}{!}{
\begin{tabular}[0.9\textwidth]{lcccccccccccc}
\toprule
Model & $\pi_{\text{LLM}}$ SR $\uparrow$ & $\pi_{\text{LLM}}$ ER $\uparrow$ & Score $\uparrow$ & AQM $\uparrow$ & StV $\downarrow$ & LEA $\uparrow$ & $|\pi_{\text{corr}}|$ $\uparrow$ & $\pi_{\text{NLP-LCS}}$ SR $\uparrow$ & $\pi_{\text{NLP-LCS}}$ StV $\downarrow$ & $\pi_{\text{NLP-LCS}}$ LEA & $\pi_{\text{VAL}}$ SR $\uparrow$ \\
\midrule
Claude Opus 4.8 & \textbf{0.54} & 0.65 & 0.72 & 0.64 & 6.57 & 0.68 & \textbf{2.28} & \textbf{0.55} & 5.74 & 0.67 & \textbf{1.0} \\
GPT 6 Sol & 0.52 & \textbf{0.73} & \textbf{0.75} & 0.68 & \textbf{4.96} & \textbf{0.73} & 2.16 & 0.52 & \textbf{4.99} & \textbf{0.69} & \textbf{1.0} \\
Gemini 3.1 Pro & 0.46 & 0.59 & 0.71 & 0.68 & 5.23 & 0.6 & 1.75 & 0.46 & 5.23 & 0.6 & \textbf{1.0} \\
Gemini 3.8 Flash & 0.41 & 0.64 & 0.54 & \textbf{0.8} & 6.36 & 0.64 & 2.22 & 0.41 & 6.36 & 0.67 & \textbf{1.0} \\
Gemma 4 12B & 0.1 & 0.23 & 0.38 & 0.4 & 11.22 & 0.32 & 0.62 & 0.12 & 9.22 & 0.37 & \textbf{1.0} \\
Gemma 4 26B & 0.08 & 0.18 & 0.27 & 0.37 & 17.04 & 0.29 & 0.71 & 0.11 & 8.33 & 0.43 & \textbf{1.0} \\
Qwen3 14B & 0.11 & 0.21 & 0.4 & 0.48 & 8.54 & 0.29 & 0.62 & 0.13 & 7.83 & 0.35 & \textbf{1.0} \\
Qwen3 8B & 0.05 & 0.14 & 0.36 & 0.42 & 10.54 & 0.23 & 0.51 & 0.08 & 8.41 & 0.31 & \textbf{1.0} \\
\bottomrule
\end{tabular}
}
\end{table}

\begin{table}[b]
\caption{Comparison of core metrics by prompt types (all domains) - optimal plan as GT}
\label{tab:core_metrics_by_task_opt}
\resizebox{\textwidth}{!}{
\begin{tabular}[0.9\textwidth]{lcccccccccccc}
\toprule
Task & $\pi_{\text{LLM}}$ SR $\uparrow$ & $\pi_{\text{LLM}}$ ER $\uparrow$ & Score $\uparrow$ & AQM $\uparrow$ & StV $\downarrow$ & LEA $\uparrow$ & $|\pi_{\text{corr}}|$ $\uparrow$ & $\pi_{\text{NLP-LCS}}$ SR $\uparrow$ & $\pi_{\text{NLP-LCS}}$ StV $\downarrow$ & $\pi_{\text{NLP-LCS}}$ LEA & $\pi_{\text{VAL}}$ SR $\uparrow$ \\
\midrule
One-Shot & 0.23 & 0.34 & 0.54 & 0.56 & 9.2 & 0.38 & 1.05 & 0.24 & 7.68 & 0.4 & \textbf{1.0} \\
PDDL & \textbf{0.58} & \textbf{0.69} & \textbf{0.68} & \textbf{0.64} & 7.07 & \textbf{0.73} & \textbf{2.32} & \textbf{0.6} & \textbf{4.73} & \textbf{0.82} & \textbf{1.0} \\
State Tracking & 0.39 & 0.57 & 0.58 & 0.61 & \textbf{6.38} & 0.62 & 2.1 & 0.4 & 6.03 & 0.64 & \textbf{1.0} \\
Zero-Shot & 0.16 & 0.39 & 0.43 & 0.54 & 9.43 & 0.41 & 0.87 & 0.16 & 7.87 & 0.37 & \textbf{1.0} \\
\bottomrule
\end{tabular}
}
\end{table}

\section*{Appendix B: Further details}
\subsection*{Running Example}
\begin{table*}[ht!]
\centering
\scriptsize
\resizebox{0.4\textwidth}{!}{
\begin{tabular}{ccccc|c}
\toprule
\textbf{Action} & (pick-up c) & (stack c a) & (stack b c) & (stack a b) & $\pi_\text{LLM}$ \\
\midrule
\textbf{Score} & 0.28 & 1.00 & 0.50 & 0.09 & 0.47 \\
\bottomrule
\end{tabular}
}
\caption{$\pi_\text{LLM}$ similarity scores per action}
\label{tab:similarity_score_by_action}
\end{table*}

Table \ref{tab:similarity_score_by_action} details the score each action in the running example achieves during the pairing process. Notably, \texttt{\small (pick-up c)} is the only \texttt{\small correct} action in our $\pi_\text{LLM}$; it receives a score of 1, as it will inevitably later be part of the longest common subsequence, making the score at least 2 for correct actions.
Tables \ref{tab:running_example} and \ref{tab:metric_comparison_for_running_example} present the information presented in the main paper for the running example.
\begin{table}[htb]
    \scriptsize
    \centering
    \resizebox{\textwidth}{!}{
    \begin{tabular}{
         *{6}{r l |@{\quad}}
              r l
        }
        \multicolumn{2}{c}{\textbf{\normalsize $\pi_\text{GT}$}} & \multicolumn{2}{c}{\textbf{\normalsize $\pi_\text{LLM}$}} & \multicolumn{2}{c}{\textbf{\footnotesize AQM}} & \multicolumn{2}{c}{\textbf{\footnotesize NP-AQM}} & \multicolumn{2}{c}{\textbf{\normalsize$\pi_\text{NLP}$}} & \multicolumn{2}{c}{\textbf{\normalsize$\pi_\text{LCS}$}} & \multicolumn{2}{c}{\textbf{\normalsize$\pi_\text{NLP-LCS}$}}\\
        \midrule
        1 & \texttt{(unstack c a)} & 1 & \texttt{(pick-up c)} & 1 & \texttt{same\_act} & 1 & \texttt{same\_act} & 1 & \texttt{(stack a b)} & 1 & \texttt{(stack c a)} & 1 & \texttt{(pick-up b)} \\
        2 & \texttt{(stack c a)} & 2 & \texttt{(stack c a)} & 2 & \texttt{correct} & 2 & \texttt{correct} & 2 & \texttt{(stack c a)} & 2 & \texttt{(stack b c)} & 2 & \texttt{(stack b c)} \\
        3 & \texttt{(pick-up b)} & 3 & \texttt{(stack b c)} & 3 & \texttt{misplaced} & 3 & \texttt{correct} & 3 & \texttt{(pick-up b)} &  &  &  &  \\
        4 & \texttt{(stack b c)} & 4 & \texttt{(stack a b)} & 4 & \texttt{diff\_act} & 4 & \texttt{diff\_act} & 4 & \texttt{(stack b c)} &  &  &  &  \\
    \end{tabular}
    }
    \caption{Consolidated running example}
    \label{tab:running_example}
\end{table}

\begin{table}[!htb]
    \footnotesize
    \centering
    \captionsetup{justification=centering}
    \resizebox{0.9\textwidth}{!}{
    \begin{tabular}{lccccccc}
    \toprule
    \textbf{Metric} & $\pi_\text{LLM}$ Score $\uparrow$ & $\pi_\text{NLP}$ Score $\uparrow$ & AQM $\uparrow$ & $\text{LLM}$ StV $\downarrow$ & $\pi_\text{NLP}$ StV $\downarrow$ & $\pi_\text{LLM}$ LEA $\uparrow$ & $\pi_\text{NLP}$ LEA $\uparrow$ \\
    \midrule
    \textbf{Value} & 0.4004 & \textbf{0.773} & 0.4688 & 3 & 1 & 0.0 & 0.0 \\
    \midrule
    \midrule
    \textbf{Metric} & $\pi_\text{LLM}$ Validity & $\pi_\text{NLP}$ Validity & $\pi_\text{LCS}$ Validity & $\pi_\text{NLP-LCS}$ Validity & $\pi_\text{NLP-LCS}$ StV $\downarrow$ & $\pi_\text{NLP-LCS}$ LEA $\uparrow$ & $\pi_\text{VAL}$ Validity \\
    \midrule
    \textbf{Value} & False & False & False & \textbf{True} & \textbf{0} & \textbf{1.0} & \textbf{True} \\
    \bottomrule
    \end{tabular}
    }
    \caption{Core metric values for running example}
    \label{tab:metric_comparison_for_running_example}
\end{table}

\subsection*{WordNet Coverage}
We check the coverage of action names in WordNet and find that 99.99\% (97647 / 97657) of action names appear in WordNet for IPC domains~\cite{mcdermott20001998} from the years 1998-2014\footnote{\url{https://github.com/potassco/pddl-instances/tree/master}} (excluding mystery domains, where the semantics of the action names do not reflect their real-world use), which are commonly used domains. To address cases where WordNet does not cover an action name, we implement a synonym map from the action name to a synonym, which is then used to calculate similarity. For example, in Blocksworld, we update the synonym map with the pair \texttt{unstack:remove}, where ``remove" is one of several dictionary synonyms for ``unstack". We choose ``remove" as it has the highest semantic similarity to ``stack", which is the negation action of ``unstack" and is therefore semantically related to it. 

\subsection*{Evaluation Metrics}
Table \ref{tab:metric_definitions} provides a description and formal definition of the metrics used in the paper.

    \begin{table}[h]
        \centering
        \scriptsize
        \resizebox{\textwidth}{!}{
        \begin{tabular}{p{2cm} p{5cm} p{6cm}}
        \toprule
        \textbf{Metric} & \textbf{Description} & \textbf{Formal Definition} \\
        \midrule
        \textbf{SR / \newline Validity}
        & Fraction of plans that reach the goal from all plans in an experiment.
        & $\text{SR} = \frac{1}{N} \sum_{i=1}^{N}[\pi_i \in \Pi_{\text{valid}}]$ \\
        \midrule
        \textbf{ER / \newline Executability}
        & Fraction of plans that follow domain constraints from all plans in an experiment.
        & 
        $\text{ER} = \frac{1}{N} \sum_{i=1}^{N}[\pi_i \in \Pi_{\text{exec}}]$
        \\
        \midrule
        \textbf{Score} 
        & Plan quality score based on similarity to $\pi^*$ and plan potential. \newline \newline 
        The score of an experiment is the mean score for all $\pi_\text{LLM}$s in that experiment.
        & 
        $\text{Score}(\pi_\text{LLM}) =
        \frac{
        \operatorname{S}(\pi_\text{LLM}, \pi_\text{GT})
        + \operatorname{S}(\pi_\text{NLP}, \pi_\text{GT})
        }{2}
        $ where $S(\pi_i, \pi_\text{GT})$ is the similarity score between $\pi_{i}$ and $\pi_\text{GT}$. \newline \newline
        $\text{Score} = \frac{1}{N} \sum_{i=1}^{N} \text{Score}(\pi_\text{LLM})$
        \\
        \midrule
        \textbf{AQM Score}
        & $\text{AQM}(\pi_\text{LLM})$ is the average quality of actions in the generated plan based on their assigned label. \newline \newline 
        The AQM of an experiment is the mean AQM score for all $\pi_\text{LLM}$s in that experiment.
        & $\text{AQM}(\pi) = \frac{1}{|\pi|} \sum_{j=1}^{|\pi|} q(a_j)$, 
        where $q(a_j)$ is the label's rank. \newline \newline \newline
        $\text{AQM} = \frac{1}{N} \sum_{i=1}^{N} \text{AQM}(\pi_\text{LLM})$ \\
        \midrule
        \textbf{LEA}
        & Propotion of the index of the last action in the plan that can be executed without violating domain constraints, out of the full plan length. \newline \newline
        The LEA of an experiment is the mean LEA for all $\pi_\text{LLM}$s in that experiment.
        & $\text{LEA}(\pi) = \frac{\max \{k \mid s_k \models \text{pre}(a_k), \, s_{k+1} = \gamma(s_k, a_k)\}}{|\pi|}$, where $\text{pre}(a)$ are the preconditions for action $a$. \newline \newline $\text{LEA} = \frac{1}{N} \sum_{i=1}^{N} \text{LEA}(\pi_\text{LLM})$ \\
        \midrule
        \textbf{StV}
        & Number of stepwise modifications needed to transform the plan into $\pi_\text{GT}$. \newline \newline
        The StV of an experiment is the mean StV for all $\pi_\text{LLM}$s in that experiment.
        & $\text{StV}(\pi) = |d(\pi, \pi^*)|$, where $d(\pi,\pi^*)$ are the operations to achieve $\pi_\text{GT}$. \newline \newline \newline $\text{StV} = \frac{1}{N} \sum_{i=1}^{N} \text{StV}(\pi_\text{LLM})$ \\
        \bottomrule
        \end{tabular}
        }
        \caption{Metrics and their formal definitions.}
        \label{tab:metric_definitions}
    \end{table}

\subsection*{Algorithm for Steps to Validity}

\begin{figure}[t]
\centering
\begin{minipage}[t]{0.48\textwidth}
    \hrule\vspace{-7pt}
    \captionof{algorithm}{Search Steps to Validity}
    \vspace{3pt}\hrule\vspace{3pt}
    \label{alg:search_steps_to_validity}
    \begin{algorithmic}[1]
    \Require $\pi_\text{LLM}$, $\pi_\text{GT}$
    \Ensure $steps$
    
    \State $steps \gets []$
    \State $repairs \gets 0$
    
    \Comment{Step 1: remove redundant actions, re-label the rest}
    \State $reduced\_plan \gets plan \setminus \{a \in plan : a.label = \textsc{Redundant}\}$
    \For{$a \in plan$ \textbf{if} $a.label = \textsc{Redundant}$}
        \State $steps.add(``remove\_action")$
    \EndFor
    \State $reduced\_plan.evaluate()$ \Comment{re-pair \& re-label against $\pi_\text{GT}$}
    
    \If{$reduced\_plan.is\_valid()$}
        \State \Return $steps$
    \EndIf
    
    \Comment{Step 2: reorder / repair remaining non-Correct actions}
    \For{$a \in reduced\_plan$}
        \If{$a.label = \textsc{Misplaced}$}
            \State $steps.add(``reorder\_action")$
            \State $increment(repairs)$
        \ElsIf{$a.label \in \{\textsc{SameAct}, \textsc{DiffAct}\}$}
            \State $steps.add(``repair\_action")$
            \State $increment(repairs)$
        \EndIf
    \EndFor
    
    \Comment{Step 3: add actions missing from $\pi_\text{GT}$, reusing repair budget where possible}
    \For{$a \in \pi_\text{GT}$}
        \If{$a \notin reduced\_plan$}
            \If{$repairs > 0$}
                \State $decrement(repairs)$ \Comment{this action's repair doubles as covering $a$}
            \Else
                \State $steps.add(``add\_action")$
            \EndIf
        \EndIf
    \EndFor
    
    \Return $steps$
    \end{algorithmic}
    \vspace{3pt}\hrule
\end{minipage}\hfill
\begin{minipage}[t]{0.48\textwidth}
    \centering
    \captionsetup{justification=centering}
    \captionof{table}{Mean absolute correlation of SR, ER, Score with remaining metrics, by model}
    \label{tab:metric_correlations_by_model}
    \begin{tabular}{lccc}
    \toprule
    Model & $\pi_{\text{LLM}}$ SR $\uparrow$ & $\pi_{\text{LLM}}$ ER $\uparrow$ & Score $\uparrow$ \\
    \midrule
    Claude Opus 4.8 & 0.64 & 0.60 & \textbf{0.66} \\
    GPT 6 Sol & 0.62 & 0.57 & \textbf{0.65} \\
    Gemini 3.1 Pro & 0.60 & \textbf{0.66} & 0.63 \\
    Gemini 3.8 Flash & 0.66 & 0.59 & \textbf{0.69} \\
    Gemma 4 12B & 0.49 & 0.44 & \textbf{0.56} \\
    Gemma 4 26B & 0.49 & 0.51 & \textbf{0.60} \\
    Qwen3 14B & 0.55 & 0.61 & \textbf{0.66} \\
    Qwen3 8B & 0.46 & 0.52 & \textbf{0.59} \\
    \bottomrule
    \end{tabular}
\end{minipage}
\end{figure}

Algorithm \ref{alg:search_steps_to_validity} describes in detail how the StV measure is computed.

\subsection*{Hungarian method for best parameter alignment}
Given a shift, the best remapping is found by solving an assignment problem. The cost matrix has one row per plan parameter $u$ and one column per candidate mapping $v$. A cell $(u,v)$ is the number of argument positions where $u$ appears in the shifted plan at the same position as $v$ in $\pi_{\text{GT}}$, among actions with the same name (and number of arguments). Maximising the total count yields a one-to-one parameter mapping, so two parameters can never be mapped to the same one. A separate matrix is solved for each type. Ties are broken in favour of keeping a parameter unchanged, and candidates with equal preliminary scores are ranked by fewer renamed parameters.
The two operations are then alternated for $r$ iterations: the best shift is chosen under the current remapping, and the best remapping under that shift. Each step is optimal given the other, but the two use different scores, so convergence to the global optimum is not guaranteed. We therefore use a fixed iteration threshold rather than a convergence criterion.

\paragraph{Hyperparameter search.}
We tuned the search on a subset of 4 models, 3 domains, and 2 prompt types, using the proportion of instances where the LLM plan was invalid and $\pi_{\text{NLP}}$ led to a valid plan as the primary metric. Each of five hyperparameters was varied one at a time with the others at their defaults: the number of candidates assessed in full ($c$), the minimum, maximum and tolerance of the shift beam ($m_{\min}$, $m_{\max}$, $\tau$), and the number of alternation iterations ($r$). Only $c$ affected the outcome; results were identical for all other values, and flat for $c \geq 3$. We therefore use $c=3$, and keep the defaults ($m_{\min}=7$, $m_{\max}=25$, $\tau=0.1$, $r=2$) for the rest. 

\begin{table}[h]
\centering
\small
\begin{tabular}{llll}
\toprule
Parameter & Meaning & Values tried & Chosen \\
\midrule
$k$ & candidates assessed in full & 3, 5, 10, 20, 40 & 3 \\
$m_{\min}$ & minimum shifts kept after preliminary scoring & fixed & 7 \\
$m_{\max}$ & maximum shifts kept after preliminary scoring & 8, 15, 25, 50 & 25 \\
$\tau$ & keep shifts within this fraction of the best score & 0.05, 0.1, 0.2, 0.4 & 0.1 \\
$r$ & alternation iterations & 0, 1, 2, 4 & 2 \\
\bottomrule
\end{tabular}
\caption{Hyperparameters of the NLP manipulation search.}
\label{tab:hp}
\end{table}

\subsection{NLP-refinement Improvement by model}
\begin{figure}[!tb]
    \centering
    \includegraphics[height=5cm]{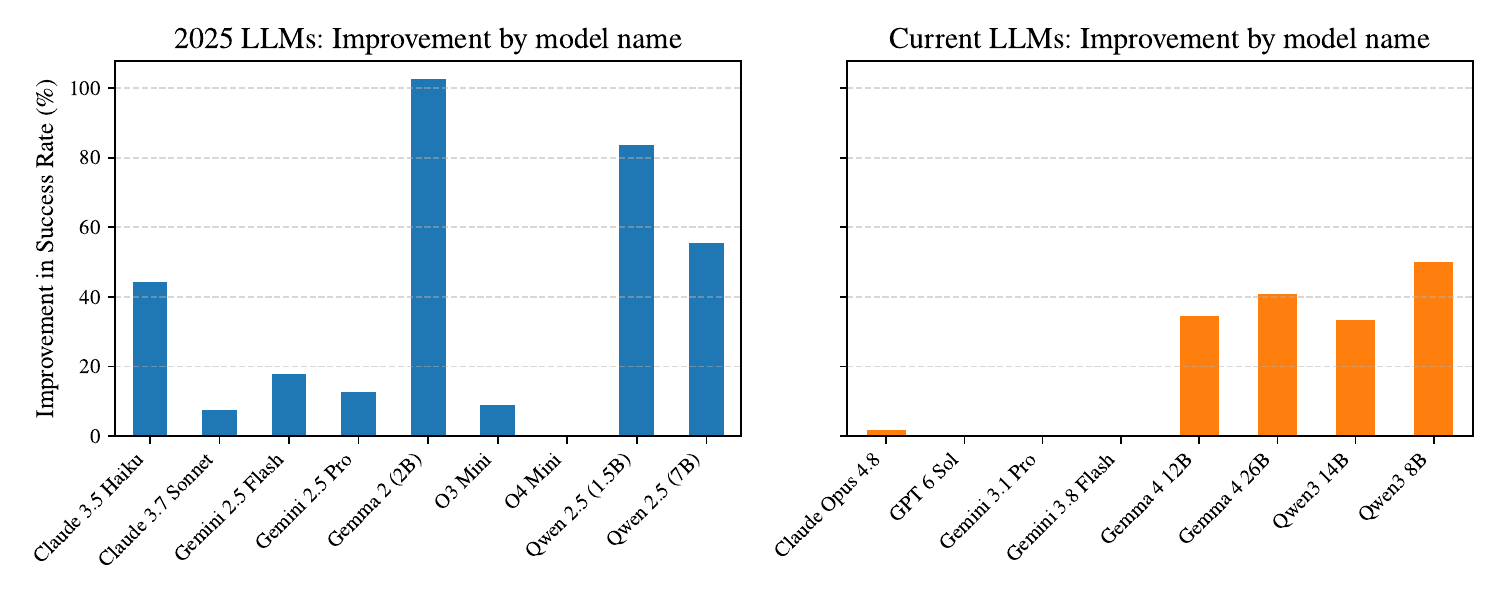}
    \caption{Validity improvement comparison across LLMs}
    \label{fig:plan_improvement_old_vs_new}
\end{figure}

\begin{figure}[!tb]
    \centering
    \includegraphics[height=4.5cm]{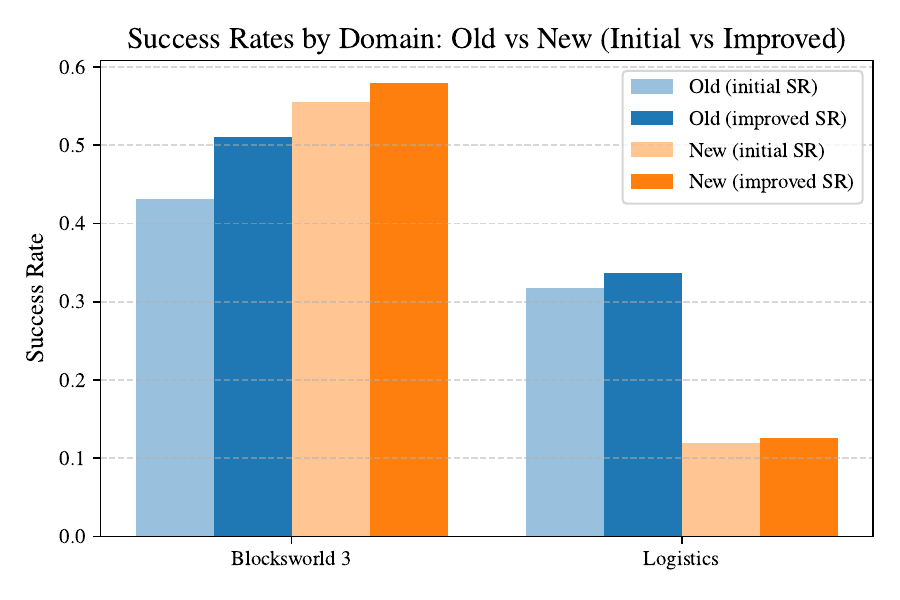}
    \caption{Validity improvement comparison across LLMs}
    \label{fig:plan_improvement_old_vs_new_2}
\end{figure}

Figures \ref{fig:plan_improvement_old_vs_new} and \ref{fig:plan_improvement_old_vs_new_2} show a comparison of validity improvement from the NLP-refinement pipeline, for current and older LLMs, for BW3 and Logistics domains. It is clear that model behaviour has changed with regards to patterns in plan generation. Table \ref{tab:core_metrics_by_domain_old} provides results for these models by domain, model, and prompt type, respectively.

\begin{table}
\caption{Comparison of core metrics by domain - 2025 LLMs}
\label{tab:core_metrics_by_domain_old}
\resizebox{\textwidth}{!}{
\begin{tabular}[0.9\textwidth]{lcccccccccccc}
\toprule
Domain & $\pi_{\text{LLM}}$ SR $\uparrow$ & $\pi_{\text{LLM}}$ ER $\uparrow$ & Score $\uparrow$ & AQM $\uparrow$ & StV $\downarrow$ & LEA $\uparrow$ & $|\pi_{\text{corr}}|$ $\uparrow$ & $\pi_{\text{NLP-LCS}}$ SR $\uparrow$ & $\pi_{\text{NLP-LCS}}$ StV $\downarrow$ & $\pi_{\text{NLP-LCS}}$ LEA & $\pi_{\text{VAL}}$ SR $\uparrow$ \\
\midrule
Blocksworld 3 & \textbf{0.43} & \textbf{0.53} & \textbf{0.7} & \textbf{0.75} & \textbf{2.98} & \textbf{0.64} & \textbf{2.74} & \textbf{0.51} & \textbf{2.2} & \textbf{0.74} & \textbf{1.0} \\
Logistics & 0.32 & 0.45 & 0.5 & 0.5 & 12.54 & 0.54 & 1.25 & 0.34 & 11.31 & 0.63 & \textbf{1.0} \\
\bottomrule
\end{tabular}
}
\end{table}

\begin{table}
\caption{Comparison of core metrics by model (all domains) - 2025 LLMs}
\label{tab:core_metrics_by_model_opt}
\resizebox{\textwidth}{!}{
\begin{tabular}[0.9\textwidth]{lcccccccccccc}
\toprule
Model & $\pi_{\text{LLM}}$ SR $\uparrow$ & $\pi_{\text{LLM}}$ ER $\uparrow$ & Score $\uparrow$ & AQM $\uparrow$ & StV $\downarrow$ & LEA $\uparrow$ & $|\pi_{\text{corr}}|$ $\uparrow$ & $\pi_{\text{NLP-LCS}}$ SR $\uparrow$ & $\pi_{\text{NLP-LCS}}$ StV $\downarrow$ & $\pi_{\text{NLP-LCS}}$ LEA & $\pi_{\text{VAL}}$ SR $\uparrow$ \\
\midrule
Claude 3.5 Haiku & 0.23 & 0.3 & 0.56 & 0.58 & 7.59 & 0.47 & 1.72 & 0.33 & 6.69 & 0.64 & \textbf{1.0} \\
Claude 3.7 Sonnet & 0.46 & 0.69 & 0.65 & 0.76 & 6.49 & 0.78 & 2.6 & 0.49 & 5.92 & 0.8 & \textbf{1.0} \\
Gemini 2.5 Flash & 0.26 & 0.5 & 0.53 & 0.66 & 8.33 & 0.66 & 1.96 & 0.3 & 7.33 & 0.7 & \textbf{1.0} \\
Gemini 2.5 Pro & 0.27 & 0.51 & 0.54 & 0.66 & 8.3 & 0.66 & 1.91 & 0.31 & 7.41 & 0.68 & \textbf{1.0} \\
Gemma 2 (2B) & 0.06 & 0.12 & 0.4 & 0.41 & 9.03 & 0.22 & 0.69 & 0.12 & 7.41 & 0.47 & \textbf{1.0} \\
O3 Mini & 0.74 & 0.74 & 0.83 & 0.75 & 6.08 & 0.8 & 3.17 & 0.8 & 5.27 & 0.85 & \textbf{1.0} \\
O4 Mini & \textbf{0.91} & \textbf{0.91} & \textbf{0.94} & \textbf{0.84} & \textbf{3.86} & \textbf{0.95} & \textbf{3.7} & \textbf{0.91} & \textbf{3.79} & \textbf{0.95} & \textbf{1.0} \\
Qwen 2.5 (1.5B) & 0.06 & 0.2 & 0.35 & 0.42 & 10.63 & 0.27 & 0.61 & 0.11 & 8.67 & 0.45 & \textbf{1.0} \\
Qwen 2.5 (7B) & 0.13 & 0.19 & 0.43 & 0.43 & 9.53 & 0.29 & 0.81 & 0.2 & 7.86 & 0.45 & \textbf{1.0} \\
\bottomrule
\end{tabular}
}
\end{table}

\begin{table}
\caption{Comparison of core metrics by prompt types (all domains) - 2025 LLMs}
\label{tab:core_metrics_by_task_opt}
\resizebox{\textwidth}{!}{
\begin{tabular}[0.9\textwidth]{lcccccccccccc}
\toprule
Task & $\pi_{\text{LLM}}$ SR $\uparrow$ & $\pi_{\text{LLM}}$ ER $\uparrow$ & Score $\uparrow$ & AQM $\uparrow$ & StV $\downarrow$ & LEA $\uparrow$ & $|\pi_{\text{corr}}|$ $\uparrow$ & $\pi_{\text{NLP-LCS}}$ SR $\uparrow$ & $\pi_{\text{NLP-LCS}}$ StV $\downarrow$ & $\pi_{\text{NLP-LCS}}$ LEA & $\pi_{\text{VAL}}$ SR $\uparrow$ \\
\midrule
One-Shot & 0.36 & 0.41 & 0.61 & 0.6 & 7.34 & 0.53 & 1.98 & 0.44 & \textbf{6.23} & 0.66 & \textbf{1.0} \\
PDDL & 0.42 & 0.44 & 0.64 & 0.61 & 7.6 & 0.55 & 2.09 & 0.47 & 6.47 & 0.67 & \textbf{1.0} \\
State Tracking & 0.26 & \textbf{0.66} & 0.5 & \textbf{0.7} & \textbf{7.22} & \textbf{0.74} & 1.83 & 0.29 & 6.97 & \textbf{0.73} & \textbf{1.0} \\
Zero-Shot & \textbf{0.47} & 0.48 & \textbf{0.65} & 0.62 & 8.1 & 0.57 & \textbf{2.23} & \textbf{0.51} & 6.6 & 0.69 & \textbf{1.0} \\
\bottomrule
\end{tabular}
}
\end{table}

\subsection{Metrics Correlation}
Tables \ref{tab:metric_correlations_by_domain}, \ref{tab:metric_correlations_by_task}, and \ref{tab:metric_correlations_by_model} present the correlation between SR, ER, Score, and the remaining metrics introduced in the paper, aggregated by domain, prompt type, and model, respectively.

\begin{table}[t]
\centering
\caption{Correlation results, grouped by domain and prompt type}
\begin{subtable}[t]{0.49\textwidth}
  \centering
  \captionsetup{justification=centering}
  \caption{Mean absolute correlation of SR, ER, Score with remaining metrics, by domain}
    \label{tab:metric_correlations_by_domain}
    \begin{tabular}{lccc}
    \toprule
    Domain & $\pi_{\text{LLM}}$ SR $\uparrow$ & $\pi_{\text{LLM}}$ ER $\uparrow$ & Score $\uparrow$ \\
    \midrule
    Blocksworld & 0.80 & 0.81 & \textbf{0.82} \\
    Blocksworld 3 & 0.71 & 0.71 & \textbf{0.72} \\
    Elevator & 0.54 & 0.52 & \textbf{0.58} \\
    Logistics & 0.67 & \textbf{0.68} & 0.66 \\
    Satellite & 0.39 & 0.40 & \textbf{0.46} \\
    \bottomrule
    \end{tabular}
\end{subtable}\hfill
\begin{subtable}[t]{0.49\textwidth}
    \captionsetup{justification=centering}
    \caption{Mean absolute correlation of SR, ER, Score with remaining metrics, by prompt type}
    \label{tab:metric_correlations_by_task}
    \begin{tabular}{lccc}
    \toprule
    Task & $\pi_{\text{LLM}}$ SR $\uparrow$ & $\pi_{\text{LLM}}$ ER $\uparrow$ & Score $\uparrow$ \\
    \midrule
    One-Shot & 0.70 & 0.68 & \textbf{0.72} \\
    PDDL & 0.55 & 0.56 & \textbf{0.58} \\
    State Tracking & 0.70 & 0.66 & \textbf{0.74} \\
    Zero-Shot & 0.57 & 0.53 & \textbf{0.60} \\
    \bottomrule
    \end{tabular}
\end{subtable}
\end{table}

\begin{table*}
\caption{Comparison of core metrics by prompt type (all domains)}
\label{tab:metric_comparison_by_task_all}
\resizebox{\textwidth}{!}{
\begin{tabular}[0.9\textwidth]{lcccccccccccc}
\toprule
Prompt Type & $\pi_{\text{LLM}}$ SR $\uparrow$ & $\pi_{\text{LLM}}$ ER $\uparrow$ & Score $\uparrow$ & AQM $\uparrow$ & StV $\downarrow$ & LEA $\uparrow$ & $|\pi_{\text{corr}}|$ $\uparrow$ & $\pi_{\text{NLP-LCS}}$ SR $\uparrow$ & $\pi_{\text{NLP-LCS}}$ StV $\downarrow$ & $\pi_{\text{NLP-LCS}}$ LEA & $\pi_{\text{VAL}}$ SR $\uparrow$ \\
\midrule
One-Shot & 0.23 & 0.33 & 0.41 & 0.43 & 10.71 & 0.38 & 0.92 & 0.24 & 8.46 & 0.42 & \textbf{1.0} \\
PDDL & \textbf{0.58} & \textbf{0.68} & \textbf{0.68} & \textbf{0.65} & 7.29 & \textbf{0.73} & 2.35 & \textbf{0.59} & \textbf{5.05} & \textbf{0.82} & \textbf{1.0} \\
State Tracking & 0.39 & 0.57 & 0.58 & 0.63 & \textbf{6.33} & 0.62 & \textbf{2.52} & 0.4 & 5.98 & 0.65 & \textbf{1.0} \\
Zero-Shot & 0.16 & 0.39 & 0.32 & 0.42 & 10.83 & 0.41 & 0.92 & 0.16 & 8.92 & 0.4 & \textbf{1.0} \\
\bottomrule
\end{tabular}
}
\end{table*}

\begin{table*}
\caption{Comparison of core metrics by prompt type - BW3}
\label{tab:metric_comparison_by_task_bw3}
\resizebox{\textwidth}{!}{
\begin{tabular}[0.9\textwidth]{lcccccccccccc}
\toprule
Prompt Type & $\pi_{\text{LLM}}$ SR $\uparrow$ & $\pi_{\text{LLM}}$ ER $\uparrow$ & Score $\uparrow$ & AQM $\uparrow$ & StV $\downarrow$ & LEA $\uparrow$ & $|\pi_{\text{corr}}|$ $\uparrow$ & $\pi_{\text{NLP-LCS}}$ SR $\uparrow$ & $\pi_{\text{NLP-LCS}}$ StV $\downarrow$ & $\pi_{\text{NLP-LCS}}$ LEA & $\pi_{\text{VAL}}$ SR $\uparrow$ \\
\midrule
One-Shot & 0.51 & 0.64 & 0.71 & 0.8 & 3.01 & 0.72 & 2.84 & 0.54 & 2.23 & 0.74 & \textbf{1.0} \\
PDDL & 0.63 & 0.76 & \textbf{0.79} & \textbf{0.88} & 2.52 & \textbf{0.82} & \textbf{3.53} & \textbf{0.67} & \textbf{1.44} & \textbf{0.87} & \textbf{1.0} \\
State Tracking & 0.49 & 0.68 & 0.69 & 0.81 & \textbf{2.31} & 0.76 & 2.81 & 0.5 & 2.12 & 0.8 & \textbf{1.0} \\
Zero-Shot & \textbf{0.64} & \textbf{0.78} & 0.73 & 0.86 & 3.45 & 0.8 & 3.28 & 0.65 & 1.67 & 0.84 & \textbf{1.0} \\
\bottomrule
\end{tabular}
}
\end{table*}

\begin{table*}
\caption{Comparison of core metrics by prompt type - BW}
\label{tab:metric_comparison_by_task_bw}
\resizebox{\textwidth}{!}{
\begin{tabular}[0.9\textwidth]{lcccccccccccc}
\toprule
Prompt Type & $\pi_{\text{LLM}}$ SR $\uparrow$ & $\pi_{\text{LLM}}$ ER $\uparrow$ & Score $\uparrow$ & AQM $\uparrow$ & StV $\downarrow$ & LEA $\uparrow$ & $|\pi_{\text{corr}}|$ $\uparrow$ & $\pi_{\text{NLP-LCS}}$ SR $\uparrow$ & $\pi_{\text{NLP-LCS}}$ StV $\downarrow$ & $\pi_{\text{NLP-LCS}}$ LEA & $\pi_{\text{VAL}}$ SR $\uparrow$ \\
\midrule
One-Shot & 0.0 & 0.0 & 0.14 & 0.18 & 14.43 & 0.01 & 0.01 & 0.02 & 8.24 & 0.12 & \textbf{1.0} \\
PDDL & 0.62 & 0.73 & 0.73 & 0.83 & 7.26 & 0.78 & 4.87 & 0.64 & 2.47 & \textbf{0.88} & \textbf{1.0} \\
State Tracking & \textbf{0.69} & \textbf{0.75} & \textbf{0.79} & \textbf{0.85} & \textbf{2.28} & \textbf{0.81} & \textbf{5.18} & \textbf{0.71} & \textbf{1.99} & 0.84 & \textbf{1.0} \\
Zero-Shot & 0.0 & 0.0 & 0.14 & 0.21 & 12.09 & 0.01 & 0.0 & 0.01 & 7.98 & 0.11 & \textbf{1.0} \\
\bottomrule
\end{tabular}
}
\end{table*}

\begin{table*}
\caption{Comparison of core metrics by prompt type - Logistics}
\label{tab:metric_comparison_by_task_logistics}
\resizebox{\textwidth}{!}{
\begin{tabular}[0.9\textwidth]{lcccccccccccc}
\toprule
Prompt Type & $\pi_{\text{LLM}}$ SR $\uparrow$ & $\pi_{\text{LLM}}$ ER $\uparrow$ & Score $\uparrow$ & AQM $\uparrow$ & StV $\downarrow$ & LEA $\uparrow$ & $|\pi_{\text{corr}}|$ $\uparrow$ & $\pi_{\text{NLP-LCS}}$ SR $\uparrow$ & $\pi_{\text{NLP-LCS}}$ StV $\downarrow$ & $\pi_{\text{NLP-LCS}}$ LEA & $\pi_{\text{VAL}}$ SR $\uparrow$ \\
\midrule
One-Shot & 0.0 & 0.02 & 0.23 & 0.23 & 16.46 & 0.02 & 0.02 & 0.0 & 14.63 & 0.03 & \textbf{1.0} \\
PDDL & \textbf{0.47} & \textbf{0.58} & \textbf{0.59} & \textbf{0.57} & \textbf{11.66} & \textbf{0.63} & \textbf{1.66} & \textbf{0.5} & \textbf{8.81} & \textbf{0.78} & \textbf{1.0} \\
State Tracking & 0.0 & 0.0 & 0.25 & 0.25 & 14.01 & 0.0 & 0.03 & 0.0 & 13.45 & 0.16 & \textbf{1.0} \\
Zero-Shot & 0.0 & 0.01 & 0.23 & 0.24 & 15.56 & 0.01 & 0.01 & 0.0 & 13.71 & 0.03 & \textbf{1.0} \\
\bottomrule
\end{tabular}
}
\end{table*}

\begin{table*}
\caption{Comparison of core metrics by prompt type - Elevator}
\label{tab:metric_comparison_by_task_elevator}
\resizebox{\textwidth}{!}{
\begin{tabular}[0.9\textwidth]{lcccccccccccc}
\toprule
Prompt Type & $\pi_{\text{LLM}}$ SR $\uparrow$ & $\pi_{\text{LLM}}$ ER $\uparrow$ & Score $\uparrow$ & AQM $\uparrow$ & StV $\downarrow$ & LEA $\uparrow$ & $|\pi_{\text{corr}}|$ $\uparrow$ & $\pi_{\text{NLP-LCS}}$ SR $\uparrow$ & $\pi_{\text{NLP-LCS}}$ StV $\downarrow$ & $\pi_{\text{NLP-LCS}}$ LEA & $\pi_{\text{VAL}}$ SR $\uparrow$ \\
\midrule
One-Shot & 0.57 & 0.61 & \textbf{0.68} & 0.53 & \textbf{4.76} & 0.67 & 1.09 & 0.57 & 4.34 & 0.77 & \textbf{1.0} \\
PDDL & \textbf{0.6} & 0.71 & \textbf{0.68} & \textbf{0.62} & 4.92 & 0.73 & 1.27 & \textbf{0.61} & \textbf{4.02} & 0.81 & \textbf{1.0} \\
State Tracking & 0.39 & \textbf{0.91} & 0.63 & 0.53 & 5.96 & \textbf{0.93} & \textbf{1.42} & 0.39 & 5.66 & \textbf{0.93} & \textbf{1.0} \\
Zero-Shot & 0.02 & 0.82 & 0.23 & 0.38 & 8.6 & 0.89 & 0.2 & 0.02 & 8.04 & 0.79 & \textbf{1.0} \\
\bottomrule
\end{tabular}
}
\end{table*}

\begin{table*}
\caption{Comparison of core metrics by model (all domains)}
\label{tab:core_metrics_by_model_all}
\resizebox{\textwidth}{!}{
\begin{tabular}[0.9\textwidth]{lcccccccccccc}
\toprule
Model & $\pi_{\text{LLM}}$ SR $\uparrow$ & $\pi_{\text{LLM}}$ ER $\uparrow$ & Score $\uparrow$ & AQM $\uparrow$ & StV $\downarrow$ & LEA $\uparrow$ & $|\pi_{\text{corr}}|$ $\uparrow$ & $\pi_{\text{NLP-LCS}}$ SR $\uparrow$ & $\pi_{\text{NLP-LCS}}$ StV $\downarrow$ & $\pi_{\text{NLP-LCS}}$ LEA & $\pi_{\text{VAL}}$ SR $\uparrow$ \\
\midrule
Claude Opus 4.8 & \textbf{0.54} & 0.65 & \textbf{0.66} & 0.59 & 7.47 & 0.68 & 2.32 & \textbf{0.55} & 6.36 & 0.68 & \textbf{1.0} \\
GPT 6 Sol & 0.52 & \textbf{0.73} & \textbf{0.66} & 0.6 & \textbf{6.23} & \textbf{0.73} & 2.3 & 0.52 & \textbf{5.74} & \textbf{0.72} & \textbf{1.0} \\
Gemini 3.1 Pro & 0.46 & 0.59 & 0.6 & 0.59 & 6.8 & 0.6 & 1.83 & 0.46 & 6.24 & 0.61 & \textbf{1.0} \\
Gemini 3.8 Flash & 0.41 & 0.64 & 0.48 & \textbf{0.68} & 6.65 & 0.64 & \textbf{2.34} & 0.41 & 6.6 & 0.68 & \textbf{1.0} \\
Gemma 4 12B & 0.1 & 0.23 & 0.35 & 0.37 & 11.7 & 0.32 & 0.59 & 0.12 & 9.63 & 0.38 & \textbf{1.0} \\
Gemma 4 26B & 0.07 & 0.18 & 0.26 & 0.36 & 17.47 & 0.28 & 0.71 & 0.1 & 8.82 & 0.44 & \textbf{1.0} \\
Qwen3 14B & 0.11 & 0.21 & 0.35 & 0.42 & 8.94 & 0.29 & 0.61 & 0.13 & 7.92 & 0.37 & \textbf{1.0} \\
Qwen3 8B & 0.05 & 0.14 & 0.31 & 0.37 & 11.34 & 0.23 & 0.46 & 0.07 & 8.8 & 0.32 & \textbf{1.0} \\
\bottomrule
\end{tabular}
}
\end{table*}

\begin{table*}
\caption{Comparison of core metrics by model - BW3}
\label{tab:core_metrics_by_model_bw3}
\resizebox{\textwidth}{!}{
\begin{tabular}[0.9\textwidth]{lcccccccccccc}
\toprule
Model & $\pi_{\text{LLM}}$ SR $\uparrow$ & $\pi_{\text{LLM}}$ ER $\uparrow$ & Score $\uparrow$ & AQM $\uparrow$ & StV $\downarrow$ & LEA $\uparrow$ & $|\pi_{\text{corr}}|$ $\uparrow$ & $\pi_{\text{NLP-LCS}}$ SR $\uparrow$ & $\pi_{\text{NLP-LCS}}$ StV $\downarrow$ & $\pi_{\text{NLP-LCS}}$ LEA & $\pi_{\text{VAL}}$ SR $\uparrow$ \\
\midrule
Claude Opus 4.8 & \textbf{1.0} & \textbf{1.0} & 0.99 & \textbf{1.0} & 0.03 & \textbf{1.0} & \textbf{5.0} & \textbf{1.0} & 0.02 & \textbf{1.0} & \textbf{1.0} \\
GPT 6 Sol & \textbf{1.0} & \textbf{1.0} & \textbf{1.0} & \textbf{1.0} & \textbf{0.0} & \textbf{1.0} & 4.95 & \textbf{1.0} & \textbf{0.0} & \textbf{1.0} & \textbf{1.0} \\
Gemini 3.1 Pro & 0.34 & 0.95 & 0.57 & 0.99 & 2.6 & 0.95 & 2.22 & 0.34 & 2.6 & \textbf{1.0} & \textbf{1.0} \\
Gemini 3.8 Flash & \textbf{1.0} & \textbf{1.0} & \textbf{1.0} & \textbf{1.0} & \textbf{0.0} & \textbf{1.0} & \textbf{5.0} & \textbf{1.0} & \textbf{0.0} & \textbf{1.0} & \textbf{1.0} \\
Gemma 4 12B & 0.16 & 0.32 & 0.56 & 0.61 & 4.11 & 0.45 & 1.37 & 0.22 & 3.8 & 0.53 & \textbf{1.0} \\
Gemma 4 26B & 0.14 & 0.36 & 0.43 & 0.62 & 8.89 & 0.51 & 1.46 & 0.2 & 3.78 & 0.61 & \textbf{1.0} \\
Qwen3 14B & 0.16 & 0.35 & 0.48 & 0.65 & 3.47 & 0.5 & 1.14 & 0.22 & 3.52 & 0.54 & \textbf{1.0} \\
Qwen3 8B & 0.09 & 0.26 & 0.46 & 0.58 & 6.46 & 0.42 & 1.02 & 0.14 & 3.96 & 0.52 & \textbf{1.0} \\
\bottomrule
\end{tabular}
}
\end{table*}

\begin{table*}
\caption{Comparison of core metrics by model - BW}
\label{tab:core_metrics_by_model_bw}
\resizebox{\textwidth}{!}{
\begin{tabular}[0.9\textwidth]{lcccccccccccc}
\toprule
Model & $\pi_{\text{LLM}}$ SR $\uparrow$ & $\pi_{\text{LLM}}$ ER $\uparrow$ & Score $\uparrow$ & AQM $\uparrow$ & StV $\downarrow$ & LEA $\uparrow$ & $|\pi_{\text{corr}}|$ $\uparrow$ & $\pi_{\text{NLP-LCS}}$ SR $\uparrow$ & $\pi_{\text{NLP-LCS}}$ StV $\downarrow$ & $\pi_{\text{NLP-LCS}}$ LEA & $\pi_{\text{VAL}}$ SR $\uparrow$ \\
\midrule
Claude Opus 4.8 & \textbf{0.55} & \textbf{0.55} & \textbf{0.63} & \textbf{0.64} & 5.24 & \textbf{0.57} & \textbf{3.99} & \textbf{0.58} & \textbf{3.57} & \textbf{0.65} & \textbf{1.0} \\
GPT 6 Sol & 0.44 & 0.44 & 0.53 & 0.54 & 7.12 & 0.45 & 3.13 & 0.46 & 4.57 & 0.52 & \textbf{1.0} \\
Gemini 3.1 Pro & 0.5 & 0.5 & 0.58 & 0.59 & 6.36 & 0.51 & 3.66 & 0.51 & 4.13 & 0.57 & \textbf{1.0} \\
Gemini 3.8 Flash & 0.32 & 0.46 & 0.44 & 0.62 & \textbf{5.12} & 0.47 & 2.41 & 0.32 & 4.96 & 0.55 & \textbf{1.0} \\
Gemma 4 12B & 0.02 & 0.03 & 0.25 & 0.29 & 12.22 & 0.12 & 0.46 & 0.05 & 7.72 & 0.2 & \textbf{1.0} \\
Gemma 4 26B & 0.03 & 0.11 & 0.2 & 0.3 & 26.94 & 0.21 & 0.83 & 0.06 & 7.18 & 0.38 & \textbf{1.0} \\
Qwen3 14B & 0.04 & 0.08 & 0.21 & 0.3 & 9.99 & 0.14 & 0.49 & 0.06 & 7.47 & 0.2 & \textbf{1.0} \\
Qwen3 8B & 0.01 & 0.04 & 0.21 & 0.28 & 11.2 & 0.08 & 0.22 & 0.04 & 7.58 & 0.2 & \textbf{1.0} \\
\bottomrule
\end{tabular}
}
\end{table*}

\begin{table*}
\caption{Comparison of core metrics by model - Logistics}
\label{tab:core_metrics_by_model_logistics}
\resizebox{\textwidth}{!}{
\begin{tabular}[0.9\textwidth]{lcccccccccccc}
\toprule
Model & $\pi_{\text{LLM}}$ SR $\uparrow$ & $\pi_{\text{LLM}}$ ER $\uparrow$ & Score $\uparrow$ & AQM $\uparrow$ & StV $\downarrow$ & LEA $\uparrow$ & $|\pi_{\text{corr}}|$ $\uparrow$ & $\pi_{\text{NLP-LCS}}$ SR $\uparrow$ & $\pi_{\text{NLP-LCS}}$ StV $\downarrow$ & $\pi_{\text{NLP-LCS}}$ LEA & $\pi_{\text{VAL}}$ SR $\uparrow$ \\
\midrule
Claude Opus 4.8 & 0.14 & 0.14 & 0.4 & 0.34 & 14.89 & 0.18 & 0.56 & 0.16 & 12.22 & 0.23 & \textbf{1.0} \\
GPT 6 Sol & \textbf{0.25} & 0.25 & \textbf{0.51} & 0.41 & \textbf{11.76} & 0.26 & \textbf{0.78} & \textbf{0.25} & 11.99 & \textbf{0.44} & \textbf{1.0} \\
Gemini 3.1 Pro & \textbf{0.25} & 0.25 & 0.48 & 0.37 & 11.85 & 0.25 & 0.7 & \textbf{0.25} & \textbf{11.73} & 0.25 & \textbf{1.0} \\
Gemini 3.8 Flash & 0.03 & \textbf{0.33} & 0.13 & \textbf{0.45} & 13.78 & \textbf{0.33} & 0.31 & 0.03 & 13.78 & 0.38 & \textbf{1.0} \\
Gemma 4 12B & 0.06 & 0.08 & 0.25 & 0.26 & 15.91 & 0.1 & 0.27 & 0.08 & 13.47 & 0.19 & \textbf{1.0} \\
Gemma 4 26B & 0.02 & 0.02 & 0.17 & 0.22 & 19.64 & 0.06 & 0.22 & 0.03 & 13.61 & 0.19 & \textbf{1.0} \\
Qwen3 14B & 0.02 & 0.04 & 0.22 & 0.25 & 13.6 & 0.04 & 0.07 & 0.02 & 12.39 & 0.09 & \textbf{1.0} \\
Qwen3 8B & 0.0 & 0.01 & 0.18 & 0.2 & 16.73 & 0.02 & 0.07 & 0.0 & 13.51 & 0.05 & \textbf{1.0} \\
\bottomrule
\end{tabular}
}
\end{table*}

\begin{table*}
\caption{Comparison of core metrics by model - Elevator}
\label{tab:core_metrics_by_model_elevator}
\resizebox{\textwidth}{!}{
\begin{tabular}[0.9\textwidth]{lcccccccccccc}
\toprule
Model & $\pi_{\text{LLM}}$ SR $\uparrow$ & $\pi_{\text{LLM}}$ ER $\uparrow$ & Score $\uparrow$ & AQM $\uparrow$ & StV $\downarrow$ & LEA $\uparrow$ & $|\pi_{\text{corr}}|$ $\uparrow$ & $\pi_{\text{NLP-LCS}}$ SR $\uparrow$ & $\pi_{\text{NLP-LCS}}$ StV $\downarrow$ & $\pi_{\text{NLP-LCS}}$ LEA & $\pi_{\text{VAL}}$ SR $\uparrow$ \\
\midrule
Claude Opus 4.8 & 0.57 & 0.84 & 0.69 & 0.56 & 4.52 & 0.87 & 1.0 & 0.58 & 4.26 & 0.89 & \textbf{1.0} \\
GPT 6 Sol & 0.53 & \textbf{0.99} & 0.66 & 0.63 & 3.52 & \textbf{0.99} & 1.34 & 0.53 & 3.45 & 0.94 & \textbf{1.0} \\
Gemini 3.1 Pro & \textbf{0.97} & \textbf{0.99} & \textbf{0.97} & \textbf{0.68} & \textbf{1.05} & \textbf{0.99} & \textbf{1.83} & \textbf{0.97} & \textbf{1.05} & \textbf{0.99} & \textbf{1.0} \\
Gemini 3.8 Flash & 0.09 & 0.64 & 0.21 & 0.64 & 7.95 & 0.64 & 0.66 & 0.09 & 7.91 & 0.63 & \textbf{1.0} \\
Gemma 4 12B & 0.24 & 0.49 & 0.41 & 0.37 & 8.61 & 0.57 & 0.57 & 0.24 & 7.7 & 0.64 & \textbf{1.0} \\
Gemma 4 26B & 0.18 & 0.25 & 0.34 & 0.32 & 11.47 & 0.38 & 0.47 & 0.18 & 7.87 & 0.58 & \textbf{1.0} \\
Qwen3 14B & 0.39 & 0.45 & 0.58 & 0.43 & 6.93 & 0.51 & 0.85 & 0.39 & 6.39 & 0.75 & \textbf{1.0} \\
Qwen3 8B & 0.14 & 0.18 & 0.41 & 0.39 & 8.13 & 0.29 & 0.33 & 0.14 & 7.79 & 0.51 & \textbf{1.0} \\
\bottomrule
\end{tabular}
}
\end{table*}

\begin{table}[ht]
\centering
\captionsetup{justification=centering}
\caption{Mean values of evaluation metrics per experiment (domain x model × prompt type) for high SR models. Bold highlights the best value per model.}
\label{tab:appendix_metrics_per_experiment_by_domain}
\resizebox{\textwidth}{!}{
\begin{tabular}{llccccccccccccc}
\toprule
\small Domain & \small Model & \small prompt type & \textbf{$\pi_{\text{LLM}}$ SR $\uparrow$} & \small \textbf{Score $\uparrow$} & \small \textbf{AQM $\uparrow$} & \small \textbf{StV $\downarrow$} & \small \textbf{$|\pi_{\text{corr}}|$ $\uparrow$} & \small \textbf{LEA $\uparrow$} & \small \textbf{$\pi_{\text{NLP}}$ SR $\uparrow$} & \small \textbf{$\pi_{\text{LCS}}$ LEA $\uparrow$} & \small \textbf{$\pi_{\text{LCS}}$ SR $\uparrow$} & \small \textbf{$\pi_{\text{NLP-LCS}}$ SR $\uparrow$} & \small \textbf{$\pi_{\text{NLP-LCS}}$ StV $\downarrow$} & \small \textbf{$\pi_{\text{VAL}}$ SR $\uparrow$} \\
\midrule
Blocksworld & Claude Opus 4.8 & One-Shot & 0.00 & 0.15 & 0.18 & 12.85 & 0.00 & 0.01 & 0.00 & 0.00 & 0.01 & 0.03 & 8.25 & \textbf{1.00} \\
Blocksworld & Claude Opus 4.8 & PDDL & 0.90 & 0.92 & 0.92 & 0.98 & \textbf{6.60} & 0.96 & 0.90 & 0.95 & 0.92 & 0.92 & 0.70 & \textbf{1.00} \\
Blocksworld & Claude Opus 4.8 & State Tracking & 0.90 & 0.94 & 0.94 & 0.74 & 6.54 & 0.92 & 0.90 & 0.92 & 0.98 & 0.95 & 0.49 & \textbf{1.00} \\
Blocksworld & Claude Opus 4.8 & Zero-Shot & 0.00 & 0.20 & 0.23 & 9.24 & 0.00 & 0.00 & 0.00 & 0.00 & 0.00 & 0.03 & 8.00 & \textbf{1.00} \\
Blocksworld 3 & Claude Opus 4.8 & One-Shot & \textbf{1.00} & 0.99 & 1.00 & 0.02 & 5.08 & \textbf{1.00} & \textbf{1.00} & \textbf{1.00} & \textbf{1.00} & \textbf{1.00} & 0.02 & \textbf{1.00} \\
Blocksworld 3 & Claude Opus 4.8 & PDDL & 0.99 & 0.98 & 0.99 & 0.05 & 5.04 & 0.99 & 0.99 & 0.99 & \textbf{1.00} & \textbf{1.00} & \textbf{0.00} & \textbf{1.00} \\
Blocksworld 3 & Claude Opus 4.8 & State Tracking & \textbf{1.00} & \textbf{1.00} & \textbf{1.00} & \textbf{0.00} & 5.00 & \textbf{1.00} & \textbf{1.00} & \textbf{1.00} & \textbf{1.00} & \textbf{1.00} & \textbf{0.00} & \textbf{1.00} \\
Blocksworld 3 & Claude Opus 4.8 & Zero-Shot & \textbf{1.00} & 0.99 & 1.00 & 0.06 & 4.87 & \textbf{1.00} & \textbf{1.00} & \textbf{1.00} & \textbf{1.00} & \textbf{1.00} & 0.06 & \textbf{1.00} \\
Elevator & Claude Opus 4.8 & One-Shot & 0.87 & 0.91 & 0.67 & 1.82 & 1.46 & 0.92 & 0.88 & 0.89 & 0.87 & 0.88 & 1.77 & \textbf{1.00} \\
Elevator & Claude Opus 4.8 & PDDL & 0.65 & 0.77 & 0.56 & 5.48 & 0.98 & 0.69 & 0.67 & 0.69 & 0.70 & 0.68 & 4.78 & \textbf{1.00} \\
Elevator & Claude Opus 4.8 & State Tracking & 0.89 & 0.89 & 0.46 & 4.74 & 1.42 & 0.95 & 0.89 & 0.89 & 0.95 & 0.89 & 4.32 & \textbf{1.00} \\
Elevator & Claude Opus 4.8 & Zero-Shot & 0.06 & 0.30 & 0.46 & 6.34 & 0.40 & \textbf{1.00} & 0.06 & 0.69 & 0.06 & 0.06 & 6.44 & \textbf{1.00} \\
Logistics & Claude Opus 4.8 & One-Shot & 0.00 & 0.29 & 0.24 & 15.57 & 0.00 & 0.00 & 0.00 & 0.00 & 0.00 & 0.00 & 14.72 & \textbf{1.00} \\
Logistics & Claude Opus 4.8 & PDDL & 0.50 & 0.68 & 0.59 & 14.26 & 1.99 & 0.63 & 0.50 & 0.59 & 0.58 & 0.57 & 8.33 & \textbf{1.00} \\
Logistics & Claude Opus 4.8 & State Tracking & 0.00 & 0.30 & 0.24 & 16.02 & 0.00 & 0.00 & 0.00 & 0.00 & 0.00 & 0.00 & 13.96 & \textbf{1.00} \\
Logistics & Claude Opus 4.8 & Zero-Shot & 0.00 & 0.28 & 0.24 & 13.32 & 0.00 & 0.00 & 0.00 & 0.00 & 0.00 & 0.00 & 12.03 & \textbf{1.00} \\
Satellite & Claude Opus 4.8 & One-Shot & 0.33 & 0.53 & 0.39 & 13.15 & 1.03 & 0.95 & 0.33 & 0.55 & 0.33 & 0.33 & 11.92 & \textbf{1.00} \\
Satellite & Claude Opus 4.8 & PDDL & 0.81 & 0.82 & 0.25 & 11.64 & 0.05 & 0.88 & 0.81 & 0.85 & 0.82 & 0.81 & 10.97 & \textbf{1.00} \\
Satellite & Claude Opus 4.8 & State Tracking & 0.27 & 0.45 & 0.30 & 17.93 & 0.25 & 0.96 & 0.27 & 0.49 & 0.27 & 0.27 & 16.45 & \textbf{1.00} \\
Satellite & Claude Opus 4.8 & Zero-Shot & 0.00 & 0.20 & 0.37 & 12.96 & 0.65 & 0.35 & 0.00 & 0.26 & 0.00 & 0.00 & 12.56 & \textbf{1.00} \\
\hline
Blocksworld & GPT 6 Sol & One-Shot & 0.00 & 0.15 & 0.18 & 12.85 & 0.00 & 0.01 & 0.00 & 0.00 & 0.01 & 0.03 & 8.25 & \textbf{1.00} \\
Blocksworld & GPT 6 Sol & PDDL & \textbf{1.00} & 0.99 & 0.99 & \textbf{0.00} & \textbf{7.27} & \textbf{1.00} & \textbf{1.00} & \textbf{1.00} & \textbf{1.00} & \textbf{1.00} & \textbf{0.00} & \textbf{1.00} \\
Blocksworld & GPT 6 Sol & State Tracking & \textbf{1.00} & 1.00 & 0.99 & \textbf{0.00} & 6.67 & \textbf{1.00} & \textbf{1.00} & \textbf{1.00} & \textbf{1.00} & \textbf{1.00} & \textbf{0.00} & \textbf{1.00} \\
Blocksworld & GPT 6 Sol & Zero-Shot & 0.00 & 0.15 & 0.18 & 12.79 & 0.00 & 0.01 & 0.00 & 0.00 & 0.01 & 0.03 & 8.19 & \textbf{1.00} \\
Blocksworld 3 & GPT 6 Sol & One-Shot & \textbf{1.00} & \textbf{1.00} & \textbf{1.00} & \textbf{0.00} & 5.00 & \textbf{1.00} & \textbf{1.00} & \textbf{1.00} & \textbf{1.00} & \textbf{1.00} & \textbf{0.00} & \textbf{1.00} \\
Blocksworld 3 & GPT 6 Sol & PDDL & \textbf{1.00} & 1.00 & \textbf{1.00} & \textbf{0.00} & 5.02 & \textbf{1.00} & \textbf{1.00} & \textbf{1.00} & \textbf{1.00} & \textbf{1.00} & \textbf{0.00} & \textbf{1.00} \\
Blocksworld 3 & GPT 6 Sol & State Tracking & \textbf{1.00} & \textbf{1.00} & \textbf{1.00} & \textbf{0.00} & 4.73 & \textbf{1.00} & \textbf{1.00} & \textbf{1.00} & \textbf{1.00} & \textbf{1.00} & \textbf{0.00} & \textbf{1.00} \\
Blocksworld 3 & GPT 6 Sol & Zero-Shot & \textbf{1.00} & \textbf{1.00} & \textbf{1.00} & \textbf{0.00} & 5.01 & \textbf{1.00} & \textbf{1.00} & \textbf{1.00} & \textbf{1.00} & \textbf{1.00} & \textbf{0.00} & \textbf{1.00} \\
Elevator & GPT 6 Sol & One-Shot & \textbf{1.00} & 0.99 & 0.74 & 0.19 & 1.96 & \textbf{1.00} & \textbf{1.00} & \textbf{1.00} & \textbf{1.00} & \textbf{1.00} & 0.19 & \textbf{1.00} \\
Elevator & GPT 6 Sol & PDDL & 0.96 & 0.94 & 0.73 & 0.64 & 1.78 & 0.96 & 0.96 & 0.97 & 0.96 & 0.96 & 0.63 & \textbf{1.00} \\
Elevator & GPT 6 Sol & State Tracking & 0.00 & 0.45 & 0.62 & 6.81 & 1.57 & \textbf{1.00} & 0.00 & 0.77 & 0.00 & 0.00 & 6.38 & \textbf{1.00} \\
Elevator & GPT 6 Sol & Zero-Shot & 0.00 & 0.20 & 0.43 & 7.46 & 0.12 & \textbf{1.00} & 0.00 & 0.70 & 0.00 & 0.00 & 7.51 & \textbf{1.00} \\
Logistics & GPT 6 Sol & One-Shot & 0.02 & 0.33 & 0.26 & 14.77 & 0.07 & 0.02 & 0.02 & 0.02 & 0.02 & 0.02 & 14.66 & \textbf{1.00} \\
Logistics & GPT 6 Sol & PDDL & 0.97 & 0.98 & 0.72 & 3.90 & 2.89 & 0.97 & 0.97 & 0.97 & 0.98 & 0.97 & 3.99 & \textbf{1.00} \\
Logistics & GPT 6 Sol & State Tracking & 0.00 & 0.37 & 0.33 & 14.32 & 0.15 & 0.02 & 0.00 & 0.02 & 0.00 & 0.00 & 14.78 & \textbf{1.00} \\
Logistics & GPT 6 Sol & Zero-Shot & 0.00 & 0.37 & 0.32 & 14.11 & 0.00 & 0.00 & 0.00 & 0.00 & 0.00 & 0.00 & 14.61 & \textbf{1.00} \\
Satellite & GPT 6 Sol & One-Shot & 0.34 & 0.60 & 0.50 & 9.33 & 1.76 & 0.96 & 0.34 & 0.63 & 0.36 & 0.34 & 8.98 & \textbf{1.00} \\
Satellite & GPT 6 Sol & PDDL & \textbf{1.00} & 0.98 & 0.32 & 1.61 & 0.39 & \textbf{1.00} & \textbf{1.00} & \textbf{1.00} & \textbf{1.00} & \textbf{1.00} & 1.61 & \textbf{1.00} \\
Satellite & GPT 6 Sol & State Tracking & 0.11 & 0.39 & 0.47 & 11.18 & 1.75 & 0.97 & 0.11 & 0.76 & 0.11 & 0.11 & 11.01 & \textbf{1.00} \\
Satellite & GPT 6 Sol & Zero-Shot & 0.10 & 0.39 & 0.48 & 11.62 & 1.69 & 0.97 & 0.10 & 0.69 & 0.10 & 0.10 & 11.34 & \textbf{1.00} \\
\hline
Blocksworld & Gemini 3.1 Pro & One-Shot & 0.00 & 0.15 & 0.18 & 12.85 & 0.00 & 0.01 & 0.00 & 0.00 & 0.01 & 0.03 & 8.25 & \textbf{1.00} \\
Blocksworld & Gemini 3.1 Pro & PDDL & \textbf{1.00} & \textbf{1.00} & 1.00 & \textbf{0.00} & \textbf{7.30} & \textbf{1.00} & \textbf{1.00} & \textbf{1.00} & \textbf{1.00} & \textbf{1.00} & \textbf{0.00} & \textbf{1.00} \\
Blocksworld & Gemini 3.1 Pro & State Tracking & \textbf{1.00} & 1.00 & 0.99 & \textbf{0.00} & 7.26 & \textbf{1.00} & \textbf{1.00} & \textbf{1.00} & \textbf{1.00} & \textbf{1.00} & \textbf{0.00} & \textbf{1.00} \\
Blocksworld & Gemini 3.1 Pro & Zero-Shot & 0.00 & 0.16 & 0.18 & 12.71 & 0.00 & 0.01 & 0.00 & 0.00 & 0.01 & 0.01 & 8.36 & \textbf{1.00} \\
Blocksworld 3 & Gemini 3.1 Pro & One-Shot & 0.34 & 0.55 & \textbf{1.00} & 2.72 & 2.29 & \textbf{1.00} & 0.34 & \textbf{1.00} & 0.34 & 0.34 & 2.72 & \textbf{1.00} \\
Blocksworld 3 & Gemini 3.1 Pro & PDDL & 0.28 & 0.57 & 0.97 & 2.62 & 2.12 & 0.87 & 0.28 & 0.87 & 0.28 & 0.28 & 2.62 & \textbf{1.00} \\
Blocksworld 3 & Gemini 3.1 Pro & State Tracking & \textbf{1.00} & \textbf{1.00} & \textbf{1.00} & \textbf{0.00} & 5.00 & \textbf{1.00} & \textbf{1.00} & \textbf{1.00} & \textbf{1.00} & \textbf{1.00} & \textbf{0.00} & \textbf{1.00} \\
Blocksworld 3 & Gemini 3.1 Pro & Zero-Shot & 0.38 & 0.55 & \textbf{1.00} & 2.62 & 2.01 & \textbf{1.00} & 0.38 & \textbf{1.00} & 0.38 & 0.38 & 2.62 & \textbf{1.00} \\
Elevator & Gemini 3.1 Pro & One-Shot & 0.99 & 0.99 & 0.71 & 0.80 & 1.91 & 0.99 & 0.99 & 0.99 & 0.99 & 0.99 & 0.80 & \textbf{1.00} \\
Elevator & Gemini 3.1 Pro & PDDL & \textbf{1.00} & 0.99 & 0.75 & 0.54 & 1.94 & \textbf{1.00} & \textbf{1.00} & \textbf{1.00} & \textbf{1.00} & \textbf{1.00} & 0.54 & \textbf{1.00} \\
Elevator & Gemini 3.1 Pro & State Tracking & \textbf{1.00} & 0.96 & 0.48 & 1.57 & 1.49 & \textbf{1.00} & \textbf{1.00} & \textbf{1.00} & \textbf{1.00} & \textbf{1.00} & 1.57 & \textbf{1.00} \\
Elevator & Gemini 3.1 Pro & Zero-Shot & 0.00 & 0.28 & 0.45 & 12.20 & 0.40 & 0.75 & 0.00 & 0.30 & 0.00 & 0.00 & 12.20 & \textbf{1.00} \\
Logistics & Gemini 3.1 Pro & One-Shot & 0.00 & 0.31 & 0.25 & 14.98 & 0.00 & 0.00 & 0.00 & 0.00 & 0.00 & 0.00 & 14.81 & \textbf{1.00} \\
Logistics & Gemini 3.1 Pro & PDDL & \textbf{1.00} & 1.00 & 0.73 & 2.74 & 2.78 & \textbf{1.00} & \textbf{1.00} & \textbf{1.00} & \textbf{1.00} & \textbf{1.00} & 2.74 & \textbf{1.00} \\
Logistics & Gemini 3.1 Pro & State Tracking & 0.00 & 0.31 & 0.25 & 14.85 & 0.00 & 0.00 & 0.00 & 0.00 & 0.00 & 0.00 & 14.71 & \textbf{1.00} \\
Logistics & Gemini 3.1 Pro & Zero-Shot & 0.00 & 0.31 & 0.25 & 14.94 & 0.00 & 0.00 & 0.00 & 0.00 & 0.00 & 0.00 & 14.73 & \textbf{1.00} \\
Satellite & Gemini 3.1 Pro & One-Shot & 0.15 & 0.47 & 0.50 & 11.57 & 0.86 & 0.49 & 0.15 & 0.33 & 0.15 & 0.15 & 11.92 & \textbf{1.00} \\
Satellite & Gemini 3.1 Pro & PDDL & 0.99 & 0.98 & 0.41 & 1.34 & 0.88 & 0.99 & 0.99 & 0.99 & 0.99 & 0.99 & 1.34 & \textbf{1.00} \\
Satellite & Gemini 3.1 Pro & State Tracking & 0.25 & 0.51 & 0.40 & 14.85 & 0.20 & \textbf{1.00} & 0.25 & 0.52 & 0.25 & 0.25 & 14.65 & \textbf{1.00} \\
Satellite & Gemini 3.1 Pro & Zero-Shot & 0.00 & 0.07 & 0.39 & 13.77 & 0.00 & 0.02 & 0.00 & 0.00 & 0.00 & 0.00 & 13.89 & \textbf{1.00} \\
\hline
Blocksworld & Gemini 3.8 Flash & One-Shot & 0.00 & 0.14 & 0.24 & 8.13 & 0.00 & 0.02 & 0.00 & 0.02 & 0.00 & 0.00 & 7.72 & \textbf{1.00} \\
Blocksworld & Gemini 3.8 Flash & PDDL & 0.20 & 0.44 & 0.91 & 4.98 & 1.89 & 0.75 & 0.20 & 0.75 & 0.20 & 0.20 & 4.98 & \textbf{1.00} \\
Blocksworld & Gemini 3.8 Flash & State Tracking & \textbf{1.00} & 0.99 & 0.99 & \textbf{0.00} & \textbf{7.28} & \textbf{1.00} & \textbf{1.00} & \textbf{1.00} & \textbf{1.00} & \textbf{1.00} & \textbf{0.00} & \textbf{1.00} \\
Blocksworld & Gemini 3.8 Flash & Zero-Shot & 0.00 & 0.13 & 0.24 & 7.92 & 0.00 & 0.01 & 0.00 & 0.00 & 0.00 & 0.00 & 7.67 & \textbf{1.00} \\
Blocksworld 3 & Gemini 3.8 Flash & One-Shot & \textbf{1.00} & \textbf{1.00} & \textbf{1.00} & \textbf{0.00} & 5.00 & \textbf{1.00} & \textbf{1.00} & \textbf{1.00} & \textbf{1.00} & \textbf{1.00} & \textbf{0.00} & \textbf{1.00} \\
Blocksworld 3 & Gemini 3.8 Flash & PDDL & \textbf{1.00} & \textbf{1.00} & \textbf{1.00} & \textbf{0.00} & 5.00 & \textbf{1.00} & \textbf{1.00} & \textbf{1.00} & \textbf{1.00} & \textbf{1.00} & \textbf{0.00} & \textbf{1.00} \\
Blocksworld 3 & Gemini 3.8 Flash & State Tracking & \textbf{1.00} & \textbf{1.00} & \textbf{1.00} & \textbf{0.00} & 5.00 & \textbf{1.00} & \textbf{1.00} & \textbf{1.00} & \textbf{1.00} & \textbf{1.00} & \textbf{0.00} & \textbf{1.00} \\
Blocksworld 3 & Gemini 3.8 Flash & Zero-Shot & \textbf{1.00} & \textbf{1.00} & \textbf{1.00} & \textbf{0.00} & 5.02 & \textbf{1.00} & \textbf{1.00} & \textbf{1.00} & \textbf{1.00} & \textbf{1.00} & \textbf{0.00} & \textbf{1.00} \\
Elevator & Gemini 3.8 Flash & One-Shot & 0.04 & 0.19 & 0.87 & 7.17 & 0.87 & \textbf{1.00} & 0.04 & \textbf{1.00} & 0.04 & 0.04 & 7.17 & \textbf{1.00} \\
Elevator & Gemini 3.8 Flash & PDDL & 0.10 & 0.23 & 0.59 & 7.79 & 0.63 & 0.56 & 0.10 & 0.56 & 0.10 & 0.10 & 7.74 & \textbf{1.00} \\
Elevator & Gemini 3.8 Flash & Zero-Shot & 0.00 & 0.02 & 0.25 & 18.67 & 0.00 & 0.33 & 0.00 & 0.33 & 0.00 & 0.00 & 18.67 & \textbf{1.00} \\
Logistics & Gemini 3.8 Flash & One-Shot & 0.00 & 0.10 & 0.31 & 14.47 & 0.06 & 0.13 & 0.00 & 0.13 & 0.00 & 0.00 & 14.47 & \textbf{1.00} \\
Logistics & Gemini 3.8 Flash & PDDL & 0.07 & 0.17 & 0.65 & 13.85 & 0.67 & 0.65 & 0.07 & 0.65 & 0.07 & 0.07 & 13.85 & \textbf{1.00} \\
Logistics & Gemini 3.8 Flash & State Tracking & 0.00 & 0.03 & 0.25 & 12.00 & 0.00 & 0.00 & 0.00 & 0.00 & 0.00 & 0.00 & 12.00 & \textbf{1.00} \\
Logistics & Gemini 3.8 Flash & Zero-Shot & 0.00 & 0.12 & 0.26 & 12.48 & 0.00 & 0.02 & 0.00 & 0.02 & 0.00 & 0.00 & 12.48 & \textbf{1.00} \\
Satellite & Gemini 3.8 Flash & One-Shot & 0.00 & 0.05 & 0.52 & 14.04 & 0.20 & 0.83 & 0.00 & 0.89 & 0.00 & 0.00 & 14.01 & \textbf{1.00} \\
Satellite & Gemini 3.8 Flash & PDDL & 0.00 & 0.06 & 0.39 & 14.31 & 0.02 & 0.32 & 0.00 & 0.32 & 0.00 & 0.00 & 14.31 & \textbf{1.00} \\
Satellite & Gemini 3.8 Flash & State Tracking & 0.00 & 0.03 & 0.25 & 16.00 & 0.00 & 0.00 & 0.00 & \textbf{1.00} & 0.00 & 0.00 & 16.00 & \textbf{1.00} \\
Satellite & Gemini 3.8 Flash & Zero-Shot & 0.00 & 0.06 & 0.57 & 13.78 & 0.28 & 0.94 & 0.00 & 0.94 & 0.00 & 0.00 & 13.78 & \textbf{1.00} \\
\bottomrule
\end{tabular}%
}
\end{table}

\begin{table}[ht]
\centering
\captionsetup{justification=centering}
\caption{Mean values of evaluation metrics per experiment (domain x model × prompt type) for low SR models. Bold highlights the best value per model.}
\label{tab:appendix_metrics_per_experiment_by_domain_2}
\resizebox{\textwidth}{!}{
\begin{tabular}{llccccccccccccc}
\toprule
\small Domain & \small Model & \small prompt type & \textbf{$\pi_{\text{LLM}}$ SR $\uparrow$} & \small \textbf{Score $\uparrow$} & \small \textbf{AQM $\uparrow$} & \small \textbf{StV $\downarrow$} & \small \textbf{$|\pi_{\text{corr}}|$ $\uparrow$} & \small \textbf{LEA $\uparrow$} & \small \textbf{$\pi_{\text{NLP}}$ SR $\uparrow$} & \small \textbf{$\pi_{\text{LCS}}$ LEA $\uparrow$} & \small \textbf{$\pi_{\text{LCS}}$ SR $\uparrow$} & \small \textbf{$\pi_{\text{NLP-LCS}}$ SR $\uparrow$} & \small \textbf{$\pi_{\text{NLP-LCS}}$ StV $\downarrow$} & \small \textbf{$\pi_{\text{VAL}}$ SR $\uparrow$} \\
\midrule
Blocksworld & Gemma 4 12B & One-Shot & 0.00 & 0.15 & 0.17 & 14.40 & 0.01 & 0.01 & 0.00 & 0.01 & 0.01 & 0.02 & 8.35 & \textbf{1.00} \\
Blocksworld & Gemma 4 12B & PDDL & 0.05 & 0.43 & 0.48 & 9.68 & 1.36 & 0.31 & 0.05 & 0.24 & 0.09 & 0.09 & 6.91 & \textbf{1.00} \\
Blocksworld & Gemma 4 12B & State Tracking & 0.08 & 0.52 & 0.59 & 5.50 & 1.58 & 0.39 & 0.08 & 0.25 & 0.17 & 0.17 & 6.08 & \textbf{1.00} \\
Blocksworld & Gemma 4 12B & Zero-Shot & 0.00 & 0.17 & 0.17 & 12.00 & 0.00 & 0.03 & 0.00 & 0.03 & 0.00 & 0.00 & 6.00 & \textbf{1.00} \\
Blocksworld 3 & Gemma 4 12B & One-Shot & 0.20 & 0.60 & 0.62 & 4.23 & 1.54 & 0.47 & 0.24 & 0.42 & 0.24 & 0.28 & 3.60 & \textbf{1.00} \\
Blocksworld 3 & Gemma 4 12B & PDDL & 0.12 & 0.56 & 0.57 & 4.00 & 1.20 & 0.34 & 0.17 & 0.33 & 0.17 & 0.24 & 4.14 & \textbf{1.00} \\
Blocksworld 3 & Gemma 4 12B & State Tracking & 0.16 & 0.55 & 0.65 & 3.60 & 1.43 & 0.54 & 0.16 & 0.50 & 0.17 & 0.17 & 3.58 & \textbf{1.00} \\
Blocksworld 3 & Gemma 4 12B & Zero-Shot & 0.00 & 0.33 & 0.39 & 9.22 & 0.00 & 0.04 & 0.00 & 0.02 & 0.00 & 0.00 & 6.33 & \textbf{1.00} \\
Elevator & Gemma 4 12B & One-Shot & 0.39 & 0.55 & 0.45 & 7.44 & 0.88 & 0.47 & 0.39 & 0.43 & 0.39 & 0.39 & 6.79 & \textbf{1.00} \\
Elevator & Gemma 4 12B & PDDL & \textbf{0.75} & \textbf{0.89} & \textbf{0.85} & \textbf{1.00} & \textbf{2.38} & \textbf{0.88} & \textbf{0.75} & \textbf{0.88} & \textbf{0.75} & \textbf{0.75} & \textbf{0.88} & \textbf{1.00} \\
Elevator & Gemma 4 12B & Zero-Shot & 0.00 & 0.19 & 0.21 & 10.84 & 0.00 & 0.67 & 0.00 & 0.20 & 0.00 & 0.00 & 9.53 & \textbf{1.00} \\
Logistics & Gemma 4 12B & One-Shot & 0.00 & 0.19 & 0.20 & 17.47 & 0.00 & 0.00 & 0.00 & 0.00 & 0.00 & 0.00 & 14.41 & \textbf{1.00} \\
Logistics & Gemma 4 12B & PDDL & 0.18 & 0.41 & 0.38 & 14.04 & 0.84 & 0.32 & 0.18 & 0.27 & 0.20 & 0.23 & 11.78 & \textbf{1.00} \\
Logistics & Gemma 4 12B & State Tracking & 0.00 & 0.17 & 0.20 & 12.79 & 0.00 & 0.00 & 0.00 & 0.00 & 0.00 & 0.00 & 10.79 & \textbf{1.00} \\
Logistics & Gemma 4 12B & Zero-Shot & 0.00 & 0.16 & 0.20 & 16.95 & 0.00 & 0.00 & 0.00 & 0.00 & 0.00 & 0.00 & 14.83 & \textbf{1.00} \\
Satellite & Gemma 4 12B & One-Shot & 0.00 & 0.26 & 0.31 & 16.04 & 0.16 & 0.18 & 0.00 & 0.07 & 0.00 & 0.00 & 13.77 & \textbf{1.00} \\
Satellite & Gemma 4 12B & PDDL & 0.08 & 0.37 & 0.31 & 18.00 & 0.23 & 0.26 & 0.08 & 0.17 & 0.08 & 0.08 & 13.19 & \textbf{1.00} \\
Satellite & Gemma 4 12B & Zero-Shot & 0.00 & 0.15 & 0.31 & 15.75 & 0.34 & 0.72 & 0.00 & 0.38 & 0.00 & 0.00 & 13.97 & \textbf{1.00} \\
\hline
Blocksworld & Gemma 4 26B & One-Shot & 0.00 & 0.10 & 0.13 & 28.35 & 0.01 & 0.01 & 0.00 & 0.01 & 0.00 & 0.02 & 8.47 & \textbf{1.00} \\
Blocksworld & Gemma 4 26B & PDDL & 0.09 & 0.33 & 0.38 & 32.07 & 1.98 & 0.28 & 0.10 & 0.26 & 0.15 & 0.15 & 5.86 & \textbf{1.00} \\
Blocksworld & Gemma 4 26B & State Tracking & 0.00 & 0.18 & 0.57 & 7.38 & 0.64 & 0.55 & 0.00 & 0.54 & 0.00 & 0.00 & 6.49 & \textbf{1.00} \\
Blocksworld & Gemma 4 26B & Zero-Shot & 0.00 & 0.04 & 0.07 & 47.67 & 0.00 & 0.00 & 0.00 & 0.01 & 0.00 & 0.00 & 8.80 & \textbf{1.00} \\
Blocksworld 3 & Gemma 4 26B & One-Shot & 0.14 & 0.44 & 0.52 & 9.95 & 1.29 & 0.35 & 0.15 & 0.32 & 0.16 & 0.18 & 4.18 & \textbf{1.00} \\
Blocksworld 3 & Gemma 4 26B & PDDL & 0.23 & \textbf{0.57} & 0.66 & 9.53 & 2.11 & 0.56 & 0.23 & 0.48 & \textbf{0.40} & 0.34 & \textbf{3.04} & \textbf{1.00} \\
Blocksworld 3 & Gemma 4 26B & State Tracking & 0.05 & 0.30 & \textbf{0.69} & \textbf{5.36} & 0.98 & 0.63 & 0.05 & 0.64 & 0.06 & 0.06 & 4.12 & \textbf{1.00} \\
Blocksworld 3 & Gemma 4 26B & Zero-Shot & \textbf{0.27} & 0.40 & 0.45 & 26.27 & 2.00 & 0.44 & \textbf{0.27} & 0.37 & 0.36 & \textbf{0.36} & 3.09 & \textbf{1.00} \\
Elevator & Gemma 4 26B & One-Shot & 0.17 & 0.28 & 0.20 & 8.71 & 0.02 & 0.32 & 0.17 & 0.30 & 0.17 & 0.17 & 6.76 & \textbf{1.00} \\
Elevator & Gemma 4 26B & PDDL & 0.22 & 0.42 & 0.42 & 12.28 & 0.82 & 0.38 & 0.22 & 0.37 & 0.22 & 0.22 & 7.84 & \textbf{1.00} \\
Elevator & Gemma 4 26B & State Tracking & 0.00 & 0.48 & 0.55 & 6.00 & \textbf{3.00} & \textbf{1.00} & 0.00 & \textbf{1.00} & 0.00 & 0.00 & 6.00 & \textbf{1.00} \\
Elevator & Gemma 4 26B & Zero-Shot & 0.00 & 0.11 & 0.20 & 20.64 & 0.00 & 0.65 & 0.00 & 0.42 & 0.00 & 0.00 & 14.18 & \textbf{1.00} \\
Logistics & Gemma 4 26B & One-Shot & 0.00 & 0.17 & 0.17 & 20.38 & 0.00 & 0.00 & 0.00 & 0.00 & 0.00 & 0.00 & 14.20 & \textbf{1.00} \\
Logistics & Gemma 4 26B & PDDL & 0.07 & 0.28 & 0.31 & 22.63 & 0.70 & 0.19 & 0.07 & 0.13 & 0.11 & 0.11 & 12.89 & \textbf{1.00} \\
Logistics & Gemma 4 26B & State Tracking & 0.00 & 0.07 & 0.24 & 13.33 & 0.00 & 0.00 & 0.00 & 0.00 & 0.00 & 0.00 & 13.29 & \textbf{1.00} \\
Logistics & Gemma 4 26B & Zero-Shot & 0.00 & 0.10 & 0.13 & 21.55 & 0.00 & 0.00 & 0.00 & 0.00 & 0.00 & 0.00 & 14.35 & \textbf{1.00} \\
Satellite & Gemma 4 26B & One-Shot & 0.00 & 0.12 & 0.25 & 19.48 & 0.18 & 0.36 & 0.00 & 0.17 & 0.00 & 0.00 & 13.87 & \textbf{1.00} \\
Satellite & Gemma 4 26B & PDDL & 0.00 & 0.14 & 0.16 & 26.14 & 0.00 & 0.00 & 0.00 & 0.00 & 0.00 & 0.00 & 11.62 & \textbf{1.00} \\
Satellite & Gemma 4 26B & Zero-Shot & 0.00 & 0.09 & 0.18 & 22.32 & 0.00 & 0.40 & 0.00 & 0.34 & 0.00 & 0.00 & 14.84 & \textbf{1.00} \\
\hline
Blocksworld & Qwen3 14B & One-Shot & 0.00 & 0.15 & 0.18 & 13.04 & 0.02 & 0.01 & 0.00 & 0.01 & 0.00 & 0.02 & 8.34 & \textbf{1.00} \\
Blocksworld & Qwen3 14B & PDDL & 0.50 & 0.56 & 0.56 & \textbf{1.00} & 1.00 & 0.50 & 0.50 & 0.50 & 0.50 & 0.50 & 1.00 & \textbf{1.00} \\
Blocksworld & Qwen3 14B & State Tracking & 0.14 & 0.49 & 0.60 & 5.82 & \textbf{2.05} & 0.55 & 0.14 & 0.44 & 0.18 & 0.18 & 5.61 & \textbf{1.00} \\
Blocksworld & Qwen3 14B & Zero-Shot & 0.00 & 0.07 & 0.26 & 7.76 & 0.00 & 0.00 & 0.00 & 0.00 & 0.00 & 0.00 & 7.65 & \textbf{1.00} \\
Blocksworld 3 & Qwen3 14B & One-Shot & 0.21 & 0.56 & 0.63 & 3.46 & 1.21 & 0.42 & 0.24 & 0.41 & 0.25 & 0.27 & 3.55 & \textbf{1.00} \\
Blocksworld 3 & Qwen3 14B & PDDL & 0.27 & 0.53 & 0.51 & 2.00 & 1.00 & 0.40 & 0.27 & 0.48 & 0.55 & 0.55 & 1.09 & \textbf{1.00} \\
Blocksworld 3 & Qwen3 14B & State Tracking & 0.21 & 0.53 & 0.72 & 3.14 & 1.57 & 0.67 & 0.22 & 0.63 & 0.22 & 0.23 & 3.37 & \textbf{1.00} \\
Blocksworld 3 & Qwen3 14B & Zero-Shot & 0.00 & 0.26 & 0.58 & 4.21 & 0.43 & 0.38 & 0.00 & 0.41 & 0.06 & 0.06 & 4.10 & \textbf{1.00} \\
Elevator & Qwen3 14B & One-Shot & 0.40 & 0.58 & 0.37 & 7.67 & 0.75 & 0.47 & 0.40 & 0.46 & 0.40 & 0.40 & 7.06 & \textbf{1.00} \\
Elevator & Qwen3 14B & PDDL & \textbf{0.62} & 0.66 & 0.70 & 1.50 & 1.88 & 0.62 & \textbf{0.62} & 0.62 & 0.62 & 0.62 & 1.50 & \textbf{1.00} \\
Elevator & Qwen3 14B & Zero-Shot & 0.00 & 0.41 & \textbf{0.92} & 2.33 & 1.17 & \textbf{1.00} & 0.00 & \textbf{1.00} & 0.00 & 0.00 & 2.33 & \textbf{1.00} \\
Logistics & Qwen3 14B & One-Shot & 0.00 & 0.20 & 0.20 & 16.91 & 0.00 & 0.00 & 0.00 & 0.00 & 0.00 & 0.00 & 14.83 & \textbf{1.00} \\
Logistics & Qwen3 14B & PDDL & 0.50 & \textbf{0.75} & 0.84 & 1.33 & 1.67 & 0.67 & 0.50 & 0.67 & \textbf{0.67} & \textbf{0.67} & \textbf{0.67} & \textbf{1.00} \\
Logistics & Qwen3 14B & State Tracking & 0.00 & 0.20 & 0.22 & 11.76 & 0.00 & 0.00 & 0.00 & 0.00 & 0.00 & 0.00 & 11.43 & \textbf{1.00} \\
Logistics & Qwen3 14B & Zero-Shot & 0.00 & 0.22 & 0.39 & 7.25 & 0.20 & 0.18 & 0.00 & 0.18 & 0.00 & 0.00 & 7.15 & \textbf{1.00} \\
Satellite & Qwen3 14B & One-Shot & 0.00 & 0.26 & 0.36 & 15.17 & 0.16 & 0.22 & 0.00 & 0.15 & 0.00 & 0.00 & 13.79 & \textbf{1.00} \\
Satellite & Qwen3 14B & PDDL & 0.00 & 0.02 & 0.25 & 16.50 & 0.00 & 0.75 & 0.00 & 0.75 & 0.00 & 0.00 & 16.50 & \textbf{1.00} \\
Satellite & Qwen3 14B & Zero-Shot & 0.00 & 0.04 & 0.39 & 12.43 & 0.14 & 0.29 & 0.00 & 0.57 & 0.00 & 0.00 & 12.43 & \textbf{1.00} \\
\hline
Blocksworld & Qwen3 8B & One-Shot & 0.00 & 0.15 & 0.18 & 12.94 & 0.01 & 0.01 & 0.00 & 0.01 & 0.01 & 0.03 & 8.29 & \textbf{1.00} \\
Blocksworld & Qwen3 8B & PDDL & 0.00 & 0.22 & 0.71 & 9.00 & 0.33 & 0.33 & 0.00 & 0.33 & 0.00 & 0.00 & 9.67 & \textbf{1.00} \\
Blocksworld & Qwen3 8B & State Tracking & 0.04 & 0.40 & 0.49 & 8.11 & 0.77 & 0.25 & 0.04 & 0.25 & 0.11 & 0.11 & 6.26 & \textbf{1.00} \\
Blocksworld & Qwen3 8B & Zero-Shot & 0.00 & 0.10 & 0.24 & 10.52 & 0.03 & 0.03 & 0.00 & 0.03 & 0.00 & 0.00 & 7.10 & \textbf{1.00} \\
Blocksworld 3 & Qwen3 8B & One-Shot & 0.15 & 0.54 & 0.61 & 3.75 & 1.27 & 0.48 & \textbf{0.18} & 0.45 & \textbf{0.17} & \textbf{0.20} & 3.78 & \textbf{1.00} \\
Blocksworld 3 & Qwen3 8B & PDDL & 0.00 & 0.28 & 0.40 & 8.25 & 0.75 & 0.21 & 0.00 & 0.21 & 0.00 & 0.00 & 4.25 & \textbf{1.00} \\
Blocksworld 3 & Qwen3 8B & State Tracking & 0.08 & 0.47 & 0.65 & 3.81 & 1.17 & 0.52 & 0.09 & 0.57 & 0.11 & 0.12 & 3.51 & \textbf{1.00} \\
Blocksworld 3 & Qwen3 8B & Zero-Shot & 0.00 & 0.28 & 0.39 & 17.06 & 0.22 & 0.12 & 0.00 & 0.17 & 0.08 & 0.06 & 5.22 & \textbf{1.00} \\
Elevator & Qwen3 8B & One-Shot & \textbf{0.17} & 0.44 & 0.38 & 7.52 & 0.33 & 0.29 & 0.17 & 0.25 & 0.17 & 0.17 & 7.13 & \textbf{1.00} \\
Elevator & Qwen3 8B & PDDL & 0.00 & 0.23 & 0.56 & 5.25 & 0.38 & 0.38 & 0.00 & 0.38 & 0.00 & 0.00 & 5.25 & \textbf{1.00} \\
Elevator & Qwen3 8B & State Tracking & 0.00 & 0.20 & 0.26 & 17.70 & 0.00 & 0.11 & 0.00 & 0.10 & 0.00 & 0.00 & 17.50 & \textbf{1.00} \\
Elevator & Qwen3 8B & Zero-Shot & 0.00 & \textbf{0.67} & \textbf{1.00} & \textbf{1.00} & \textbf{2.00} & \textbf{1.00} & 0.00 & \textbf{1.00} & 0.00 & 0.00 & \textbf{1.00} & \textbf{1.00} \\
Logistics & Qwen3 8B & One-Shot & 0.00 & 0.20 & 0.20 & 17.47 & 0.00 & 0.00 & 0.00 & 0.00 & 0.00 & 0.00 & 14.86 & \textbf{1.00} \\
Logistics & Qwen3 8B & PDDL & 0.11 & 0.40 & 0.48 & 13.22 & 1.56 & 0.43 & 0.11 & 0.35 & 0.11 & 0.11 & 12.22 & \textbf{1.00} \\
Logistics & Qwen3 8B & State Tracking & 0.00 & 0.18 & 0.20 & 12.48 & 0.00 & 0.00 & 0.00 & 0.00 & 0.00 & 0.00 & 11.30 & \textbf{1.00} \\
Logistics & Qwen3 8B & Zero-Shot & 0.00 & 0.12 & 0.17 & 19.50 & 0.00 & 0.01 & 0.00 & 0.01 & 0.00 & 0.00 & 13.18 & \textbf{1.00} \\
Satellite & Qwen3 8B & One-Shot & 0.01 & 0.30 & 0.33 & 15.60 & 0.38 & 0.29 & 0.01 & 0.12 & 0.01 & 0.01 & 13.60 & \textbf{1.00} \\
Satellite & Qwen3 8B & PDDL & 0.00 & 0.11 & 0.23 & 15.60 & 0.20 & 0.40 & 0.00 & 0.40 & 0.00 & 0.00 & 14.40 & \textbf{1.00} \\
Satellite & Qwen3 8B & Zero-Shot & 0.00 & 0.15 & 0.25 & 16.53 & 1.00 & 0.63 & 0.00 & 0.36 & 0.00 & 0.00 & 15.07 & \textbf{1.00} \\
\hline
\bottomrule
\end{tabular}%
}
\end{table}

\end{document}